\documentclass[twoside,twocolumn,9pt]{article}
\usepackage{extsizes}
\usepackage[super,sort&compress,comma]{natbib} 
\usepackage[version=3]{mhchem}
\usepackage[left=1.5cm, right=1.5cm, top=1.785cm, bottom=2.0cm]{geometry}
\usepackage{balance}
\usepackage{mathptmx}
\usepackage{amsmath}
\usepackage{sectsty}
\usepackage{graphicx} 
\usepackage{lastpage}
\usepackage[format=plain,justification=justified,singlelinecheck=false,font={stretch=1.125,small,sf},labelfont=bf,labelsep=space]{caption}
\usepackage{float}
\usepackage{fancyhdr}
\usepackage{fnpos}
\usepackage[english]{babel}
\addto{\captionsenglish}{%
  
}
\usepackage{bm}
\usepackage{array}
\usepackage{droidsans}
\usepackage{charter}
\usepackage[T1]{fontenc}
\usepackage[usenames,dvipsnames]{xcolor}
\usepackage{setspace}
\usepackage[compact]{titlesec}
\usepackage{hyperref}

\usepackage{epstopdf}

\definecolor{cream}{RGB}{222,217,201}

\begin{document}

\pagestyle{fancy}
\thispagestyle{plain}
\fancypagestyle{plain}{
\renewcommand{\headrulewidth}{0pt}
}

\makeFNbottom
\makeatletter
\renewcommand\LARGE{\@setfontsize\LARGE{15pt}{17}}
\renewcommand\Large{\@setfontsize\Large{12pt}{14}}
\renewcommand\large{\@setfontsize\large{10pt}{12}}
\renewcommand\footnotesize{\@setfontsize\footnotesize{7pt}{10}}
\makeatother

\renewcommand{\thefootnote}{\fnsymbol{footnote}}
\renewcommand\footnoterule{\vspace*{1pt}%
\color{cream}\hrule width 3.5in height 0.4pt \color{black}\vspace*{5pt}} 
\setcounter{secnumdepth}{5}

\makeatletter 
\renewcommand\@biblabel[1]{#1}            
\renewcommand\@makefntext[1]%
{\noindent\makebox[0pt][r]{\@thefnmark\,}#1}
\makeatother 
\renewcommand{\figurename}{\small{Fig.}~}
\sectionfont{\sffamily\Large}
\subsectionfont{\normalsize}
\subsubsectionfont{\bf}
\setstretch{1.125} 
\setlength{\skip\footins}{0.8cm}
\setlength{\footnotesep}{0.25cm}
\setlength{\jot}{10pt}
\titlespacing*{\section}{0pt}{4pt}{4pt}
\titlespacing*{\subsection}{0pt}{15pt}{1pt}

\fancyfoot{}
\fancyfoot[LO,RE]{\vspace{-7.1pt}\includegraphics[height=9pt]{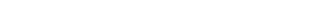}}
\fancyfoot[CO]{\vspace{-7.1pt}\hspace{13.2cm}\includegraphics{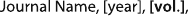}}
\fancyfoot[CE]{\vspace{-7.2pt}\hspace{-14.2cm}\includegraphics{head_foot/RF}}
\fancyfoot[RO]{\footnotesize{\sffamily{1--\pageref{LastPage} ~\textbar  \hspace{2pt}\thepage}}}
\fancyfoot[LE]{\footnotesize{\sffamily{\thepage~\textbar\hspace{3.45cm} 1--\pageref{LastPage}}}}
\fancyhead{}
\renewcommand{\headrulewidth}{0pt} 
\renewcommand{\footrulewidth}{0pt}
\setlength{\arrayrulewidth}{1pt}
\setlength{\columnsep}{6.5mm}
\setlength\bibsep{1pt}

\makeatletter 
\newlength{\figrulesep} 
\setlength{\figrulesep}{0.5\textfloatsep} 

\newcommand{\topfigrule}{\vspace*{-1pt}%
\noindent{\color{cream}\rule[-\figrulesep]{\columnwidth}{1.5pt}} }

\newcommand{\botfigrule}{\vspace*{-2pt}%
\noindent{\color{cream}\rule[\figrulesep]{\columnwidth}{1.5pt}} }

\newcommand{\dblfigrule}{\vspace*{-1pt}%
\noindent{\color{cream}\rule[-\figrulesep]{\textwidth}{1.5pt}} }

\makeatother

\twocolumn[
  \begin{@twocolumnfalse}
{\includegraphics[height=30pt]{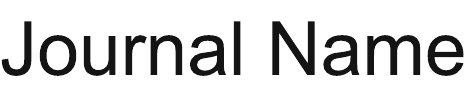}\hfill\raisebox{0pt}[0pt][0pt]{\includegraphics[height=55pt]{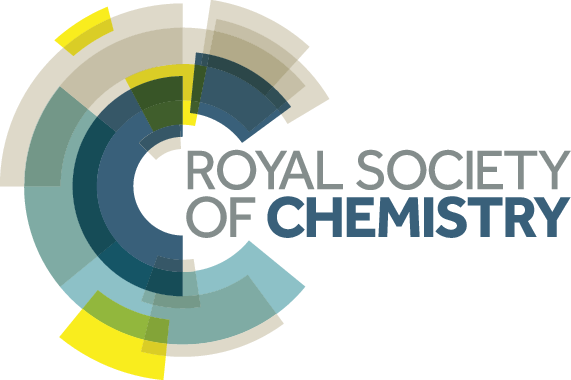}}\\[1ex]
\includegraphics[width=18.5cm]{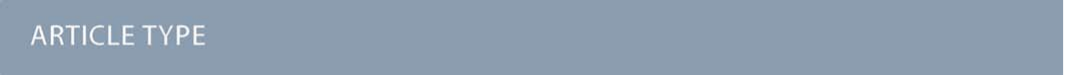}}\par
\vspace{1em}
\sffamily
\begin{tabular}{m{4.5cm} p{13.5cm} }

\includegraphics{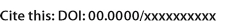} & \noindent\LARGE{\textbf{Synergistic Fusion of Multi-Source Knowledge via Evidence Theory for High-Entropy Alloy Discovery}} \\
\vspace{0.3cm} & \vspace{0.3cm} \\

 & \noindent\large{Minh-Quyet Ha,\textit{$^{a}$} Dinh-Khiet Le,\textit{$^{a}$} Duc-Anh Dao,\textit{$^{a}$} Tien-Sinh Vu,\textit{$^{a}$} Duong-Nguyen Nguyen,\textit{$^{a}$} Viet-Cuong Nguyen,\textit{$^{b}$} Hiori Kino,\textit{$^{c}$} Van-Nam Huynh,\textit{$^{a}$} and Hieu-Chi Dam$^{\ast}$\textit{$^{a}$}} \\

\includegraphics{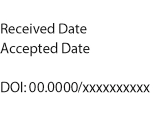} & \noindent\normalsize{Discovering novel high-entropy alloys (HEAs) with desirable properties is challenging due to the vast compositional space and complex phase formation mechanisms. Efficient exploration of this space requires a strategic approach that integrates heterogeneous knowledge sources. Here, we propose a framework that systematically combines knowledge extracted from computational material datasets with domain knowledge distilled from scientific literature using large language models (LLMs). A central feature of this approach is the explicit consideration of element substitutability, identifying chemically similar elements that can be interchanged to potentially stabilize desired HEAs. Dempster-Shafer theory, a mathematical framework for reasoning under uncertainty, is employed to model and combine substitutabilities based on aggregated evidence from multiple sources. The framework predicts the phase stability of candidate HEA compositions and is systematically evaluated on both quaternary alloy systems, demonstrating superior performance compared to baseline machine learning models and methods reliant on single-source evidence in cross-validation experiments. By leveraging multi-source knowledge, the framework retains robust predictive power even when key elements are absent from the training data, underscoring its potential for knowledge transfer and extrapolation. Furthermore, the enhanced interpretability of the methodology offers insights into the fundamental factors governing HEA formation. Overall, this work provides a promising strategy for accelerating HEA discovery by integrating computational and textual knowledge sources, enabling efficient exploration of vast compositional spaces with improved generalization and interpretability.} \\

\end{tabular}

 \end{@twocolumnfalse} \vspace{0.6cm}

  ]

\renewcommand*\rmdefault{bch}\normalfont\upshape
\rmfamily
\section*{}
\vspace{-1cm}


\footnotetext{\textit{$^{a}$~Japan Advanced Institute of Science and Technology, 1-1 Asahidai, Nomi, Ishikawa 923-1292, Japan}}
\footnotetext{\textit{$^{b}$~HPC SYSTEMS Inc., Tokyo, Japan}}
\footnotetext{\textit{$^{c}$~Research and Services Division of Materials Data and Integrated System, National Institute for Materials Science, 1-2-1 Sengen, Tsukuba, Ibaraki 305-0044, Japan}}


\section{Introduction}

High-entropy alloys (HEAs), also referred to as multiprincipal element alloys (MPEAs), have generated considerable interest because of their exceptional mechanical properties, thermal stability, and corrosion resistance~\cite{Yeh2004,CANTOR2004213,Senkov2015}. These alloys typically consist of five or more principal elements in near-equiatomic ratios, creating a high configurational entropy of mixing that stabilizes single-phase solid solutions~\cite{Rickman2019, Tsai2014}. Throughout this work, we use the term “HEA” to describe alloys composed of multiple equiatomically combined elements that form disordered solid-solution phases. Despite the promise of HEAs, identifying stable single-phase compositions remains a significant challenge due to the vastness of the design space and the complex interplay among mixing entropy, enthalpy, atomic size differences, and electronic structure.

\begin{figure}[t]
\centering
\includegraphics[width=1.0\columnwidth]{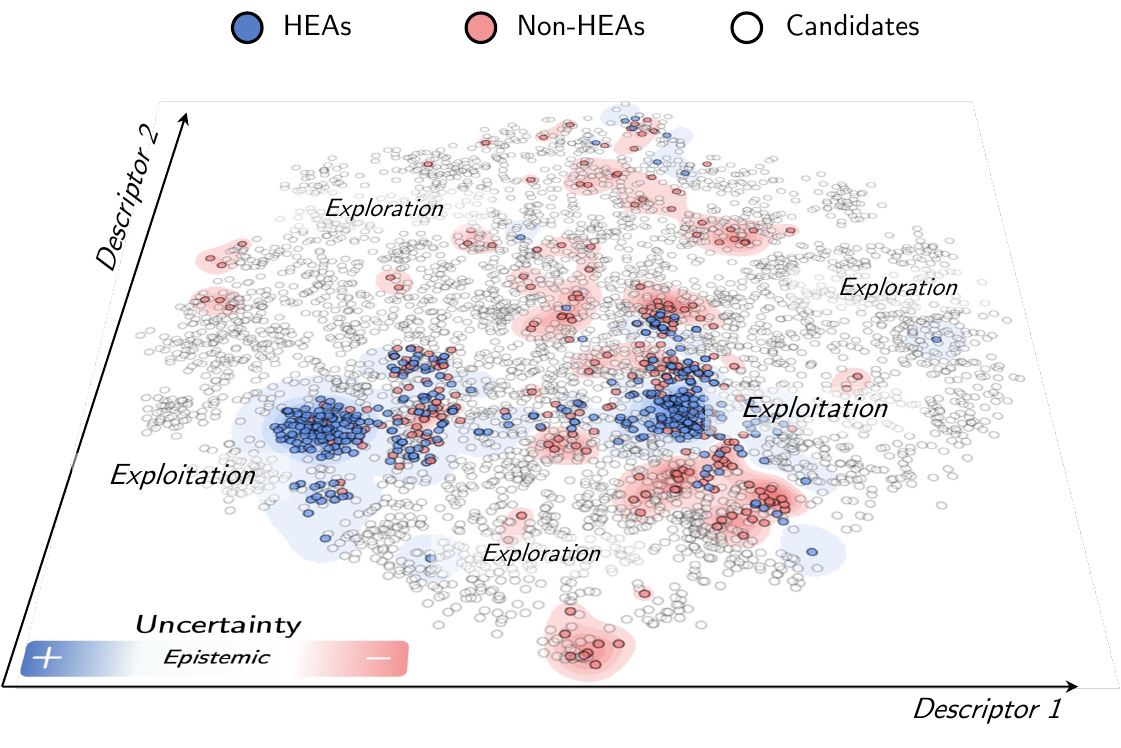}
\caption{The material space illustrates decision-making scenarios for exploitation and exploration criteria in High-Entropy Alloy (HEA) discovery. Colored regions represent familiar regions of the HEA space where sufficient data is available, while white regions represent novel, unexplored areas where data is sparse or absent.}
\label{fgr:dm_scenarios}
\end{figure}

A useful lens for understanding this challenge is a decision-making framework, wherein researchers must strike a balance between \emph{exploitation} and \emph{exploration}~\cite{Gaurav2024,Ghorbani2024}, as illustrated in Fig.~\ref{fgr:dm_scenarios}. Exploitation refers to investigating regions where sufficient data exist to predict properties with reasonable certainty, enabling incremental improvements on known alloys. Exploration, on the other hand, involves probing novel regions of the design space, where data are sparse or nonexistent and uncertainties are higher. Although exploration bears more risk, it also holds the greatest potential for discovering fundamentally new alloys with transformative properties. Finding the right balance between these two strategies is critical for accelerating HEA development.

Data-driven methods have emerged as transformative tools to guide these decisions by processing large datasets and streamlining the search for promising HEAs~\cite{Tsai2016,Tsai2019,HUANG2019225,Ziyuan2022,Josiah2024}. High-throughput techniques, such as CALPHAD~\cite{Alman2013,Senkov2015} and AFLOW~\cite{Curtarolo2012,ZHANG20141,Toher2024}, along with machine learning (ML)\cite{Gus2021}, have significantly reduced the time and cost of evaluating candidate compositions. Furthermore, approaches like active learning and Bayesian optimization can adaptively select experiments or simulations that yield the most informative data, enhancing overall efficiency~\cite{Nguyen2023,Gaurav2024,Ghorbani2024}. Yet a core limitation persists: conventional ML models excel at \emph{interpolation}—predicting outcomes for compositions similar to those in their training sets—while often struggling to \emph{extrapolate} to novel alloy systems~\cite{Ha2021}. HEA discovery inherently requires venturing beyond well-established knowledge, highlighting a fundamental tension between relying on existing data and exploring unexplored territory.

A critical aspect of managing this exploration-exploitation balance is uncertainty quantification, which can be viewed in two primary forms. \emph{Epistemic uncertainty} stems from incomplete or sparse data, whereas \emph{aleatoric uncertainty} corresponds to intrinsic variability in the system~\cite{Hullermeier2021}. Traditional methods like Bayesian neural networks, Gaussian processes, and Monte Carlo dropout can help quantify these uncertainties, but often struggle under conditions of conflicting~\cite{Hullermeier2008} or minimal data—commonly encountered in early-stage materials discovery~\cite{George2019,Konno2021}. Dempster-Shafer Theory~\cite{dempster1968, Shafer1976, denoeux20b} (referred to as the evidence theory) offers a useful alternative because it represents uncertainty through belief functions, allowing partial belief across multiple hypotheses. This enables a more nuanced treatment of both epistemic and aleatoric uncertainties, guiding researchers toward regions of the composition space where data collection can reduce uncertainty~\cite{Ha2021,Ha2023} or, conversely, toward areas with intrinsically high variability that might still hold valuable discoveries~\cite{Ton2021,Ha2023}.

Another complementary strategy for addressing high epistemic uncertainty is to leverage expert feedback. Domain specialists draw on experience accumulated over multiple studies and contexts, offering insights beyond the scope of any single dataset~\cite{Knapp2022,David2024,Liu2024}. These expert perspectives are especially valuable in uncharted compositional spaces where predictive models cannot rely on extensive prior data, and where interdisciplinary knowledge can expedite identifying promising novel alloys.

In this study, we propose a hybrid framework that combines AI-driven approaches with expert insights to improve decision-making in HEA discovery. Central to our methodology is the \emph{elemental substitution} principle, a familiar concept in alloy design wherein one or more elements are swapped for chemically similar counterparts while preserving or enhancing target properties. We treat each pair of observed alloys as evidence of which elemental combinations contribute to these properties. Subsequently, we amalgamate data-driven evidence with expert feedback obtained through large language models (LLMs), specifically employing GPT-4o, to derive additional insights regarding potential substitutions. By integrating and weighting evidence through a Dempster-Shafer–based inference mechanism, our framework systematically manages both epistemic and aleatoric uncertainties, enabling more informed decisions about where to explore next. This approach aims not only to refine predictions in well-established composition spaces, but also to promote the discovery of truly novel HEAs in data-scarce, where it compensates for the lack of empirical observations, or high-uncertainty regions, ultimately advancing the frontier of materials science research.

\begin{figure*}[t]
\centering
\includegraphics[width=1.0\textwidth]{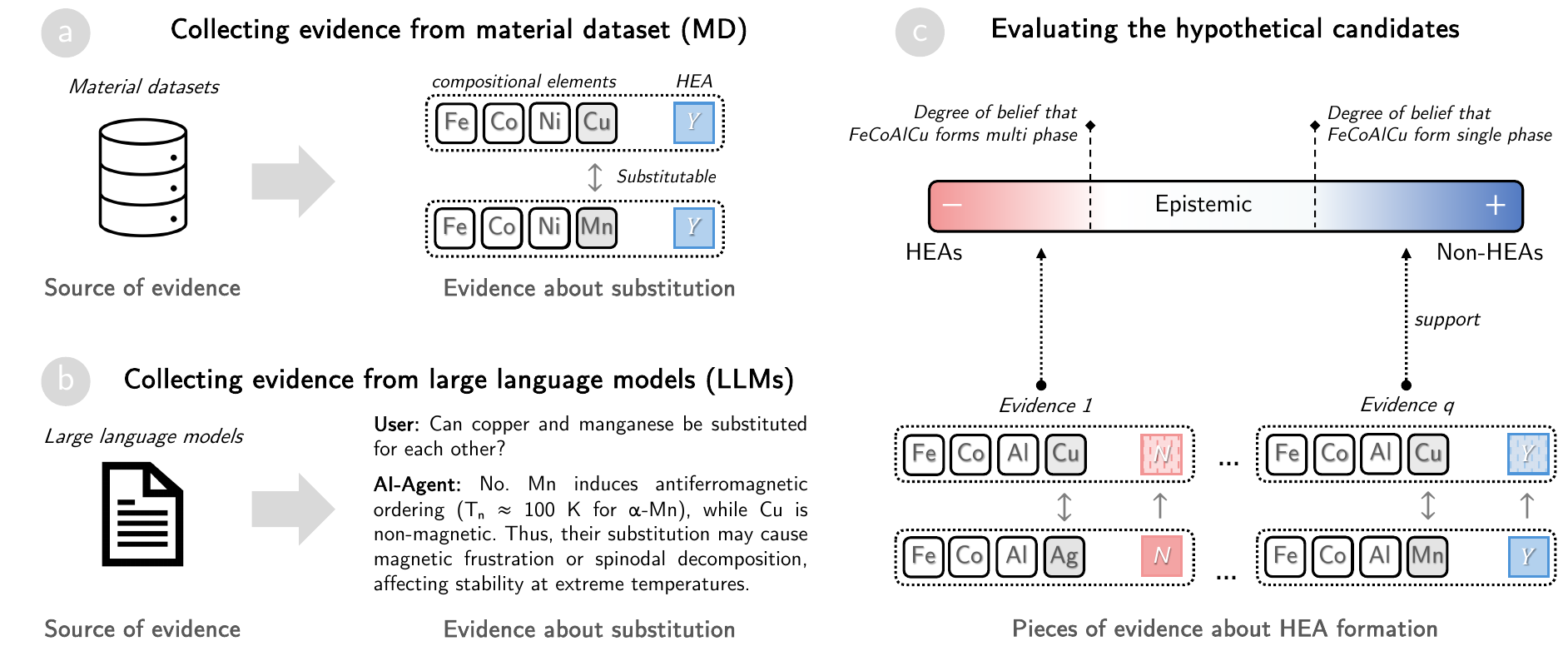}
\caption{Workflow illustration of the proposed method for evaluating hypothetical candidates forming high-entropy alloy (HEA) phases. (a–b) Schematic outlining the collection of substitutability evidence from a single material dataset (MD) and large language models (LLMs). (c) Schematic for assessing the properties of hypothetical candidates using aggregated evidence derived from substitution-based methods.}
\label{fig:hybrid_framework}
\end{figure*}

\section{Methodology}

Each alloy $A$ in the dataset $\mathcal{D}$ is represented by its constituent elements. The property of interest $y_A$ for any alloy $A$ can be either $HEA$ (indicating the presence of a high-entropy phase) or $\neg{HEA}$ (indicating no such phase). To determine elemental substitutability, we measure the similarity between different element combinations by adapting \emph{evidence theory}, which models and aggregates diverse pieces of evidence obtained from $\mathcal{D}$.

Similarity between objects can manifest in various forms~\cite{Tversky1978}, such as pairwise ratings, object sorting, communal associations, substitutability, and correlation. In this work, we focus on the \emph{solid-solution formability} of element combinations and quantify their similarity in terms of the substitutability of elements.

\subsection{Transforming Materials Data to Substitutability Evidence}
\label{subsec:md-evidence}

Consider two alloys $A_i$ and $A_j$ in $\mathcal{D}$ that share at least one common element. This nondisjoint pair of alloys provides evidence regarding the substitutability between the element combinations\[C_t = A_i \setminus (A_i \cap A_j) \quad \text{and} \quad C_v = A_j \setminus (A_i \cap A_j).\] The intersection $A_i \cap A_j$ acts as the \emph{context} for measuring similarity. If $y_{A_i}$ and $y_{A_j}$ agree (i.e., both are $HEA$ or both are $\neg{HEA}$), we infer that $C_t$ and $C_v$ are substitutable. Otherwise, they are non-substitutable.

To formally capture the evidence for similarity, we define a \emph{frame of discernment}~\cite{Shafer1976} $\Omega_{sim} = \{\mathrm{similar},\,\mathrm{dissimilar}\}$, which encompasses all possible outcomes. The evidence from $A_i$ and $A_j$ is then represented by a \emph{mass function} (or \emph{basic probability assignment}) $m^{C_t,C_v}_{A_i,A_j}$. This mass function assigns nonzero probability to the nonempty subsets of $\Omega_{sim}$, as follows.

\begin{equation}
m^{C_t,C_v}_{A_i,A_j}(\{\mathrm{similar}\}) =
  \begin{cases}
    \alpha,       & \text{if } y_{A_i} = y_{A_j},\\
    0,            & \text{otherwise},
  \end{cases}
\end{equation}

\begin{equation}
m^{C_t,C_v}_{A_i,A_j}(\{\mathrm{dissimilar}\}) =
  \begin{cases}
    \alpha,       & \text{if } y_{A_i} \neq y_{A_j},\\
    0,            & \text{otherwise},
  \end{cases}
\end{equation}

\begin{equation}
m^{C_t,C_v}_{A_i,A_j}(\Omega_{sim}) = 1 - \alpha.
\end{equation}

Here, the parameter $0 < \alpha < 1$ is determined through an exhaustive search for the best cross-validation performance, \textcolor{blue}{as shown in Supplementary Section IV}. Intuitively, $m^{C_t,C_v}_{A_i,A_j}(\{\mathrm{similar}\})$ and $m^{C_t,C_v}_{A_i,A_j}(\{\mathrm{dissimilar}\})$ capture the extent to which alloys $A_i$ and $A_j$ support substitutability or non-substitutability of $C_t$ and $C_v$, whereas $m^{C_t,C_v}_{A_i,A_j}(\Omega_{sim})$ encodes the epistemic uncertainty (i.e., lack of definitive information). The probabilities assigned to these three subsets of $\Omega_{sim}$ must sum to 1.

Suppose we collect $q$ pieces of evidence from $\mathcal{D}$ to compare $C_t$ and $C_v$. Each piece of evidence corresponds to a pair of alloys that generates a mass function $m_i^{C_t,C_v}$. We then combine these $q$ mass functions via \emph{Dempster's rule of combination}~\cite{dempster1968} to obtain a joint mass function $m^{C_t,C_v}_{\mathcal{D}}$. 
\begin{equation}
m^{C_t,C_v}_{\mathcal{D}}(\omega) \;=\; 
\Bigl(m_1^{C_t,C_v} \oplus m_2^{C_t,C_v} \oplus \dots \oplus m_q^{C_t,C_v}\Bigr)(\omega),
\end{equation}
for each $\omega \subseteq \Omega_{sim}$, $\omega \neq \emptyset$.  $\oplus$ indicate for the Dempster's rule of combinations, as shown in \textcolor{blue}{Supplementary Section II}. If no relevant evidence is available, then $m^{C_t,C_v}_{\mathcal{D}}$ is initialized with a mass of 1 on $\{\mathrm{similar}, \mathrm{dissimilar}\}$, indicating total uncertainty.

\begin{table*}[ht]
    \centering
    \caption{Possible outcomes from one of LLMs for each domain-specific criteria and the corresponding mass functions 
    $m_{\mathrm{LLMs}}^{C_t,C_v}(\{\mathrm{similar}\})$, 
    $m_{\mathrm{LLMs}}^{C_t,C_v}(\{\mathrm{dissimilar}\})$, and 
    $m_{\mathrm{LLMs}}^{C_t,C_v}(\{\mathrm{similar}, \mathrm{dissimilar}\})$. 
    Here, $0 < \beta < 1$ indicates our confidence in GPT response.}
    \label{tab:gpt-outcomes}
    \begin{tabular}{ccccccc}
    \hline
    \textbf{Question 1} & \textbf{Question 2} & $m_{\mathrm{LLMs}}^{C_t,C_v}(\{\mathrm{similar}\})$ & $m_{\mathrm{LLMs}}^{C_t,C_v}(\{\mathrm{dissimilar}\})$ & $m_{\mathrm{LLMs}}^{C_t,C_v}(\Omega_{sim})$ & \textbf{Interpretation}\\
    \hline
    \texttt{No}  & -- & $0$ & $0$ & $1$ & No knowledge \\
    \texttt{Yes} & \texttt{High} & $\beta$ & $0$ & $1-\beta$ & High substitutability \\
    \texttt{Yes} & \texttt{Medium} & $\beta /2$ & $\beta /2$ & $1-\beta$ & Medium substitutability \\
    \texttt{Yes} & \texttt{Low} & $0$ & $\beta$ & $1-\beta$ & Low substitutability \\
    \hline
    \end{tabular}
\end{table*}

\subsection{Transforming Domain Knowledge to Substitutability Evidence}
\label{subsec:llm-evidence}

In the previous section, we established evidence collection from material datasets. Here, we focus on evidence derived from domain knowledge, leveraging large language models (LLMs) to distill insights from a vast corpus of scientific literature. Specifically, we utilize GPT-4o to gather information on element substitutability. These model evaluates the substitutability of element pairs based on the perspective of an expert in a specific domain. To ensure the reliability of the results, we design a two-step prompt structure:
\begin{itemize}
\item \textbf{Question 1:} Do you possess sufficient knowledge/data to assess the substitutability of elements $C_t$ and $C_v$  regarding \textit{[domain knowledge]}?
\item \textbf{Question 2:} If the answer for the first question is \texttt{Yes}, rate their substitutability as \texttt{High}, \texttt{Medium}, or \texttt{Low}.
\end{itemize}

Detailed prompts used for each LLM evaluation are provided in \textcolor{blue}{Supplementary File 1}. The assumption relies on the premise that, given clear and structured prompts, GPT-4o can simulate expert reasoning across multiple scientific domains. This is made possible by its extensive training on scientific literature, which equips it to provide contextually relevant, domain-specific feedback aligned with HEA discovery challenges.

To guide the extraction of domain knowledge, we utilize five key scientific domains—\textit{Corrosion Science}, \textit{Materials Mechanics}, \textit{Metallurgy}, \textit{Solid-State Physics}, and \textit{Materials Science}—due to their significant contributions to understanding and optimizing high-entropy alloys (HEAs)~\cite{Tsai2014}. Each domain provides critical insights into different aspects of alloy design:

\begin{itemize}
\item \textbf{Corrosion Science}: Focuses on chemical degradation and protection strategies, essential for ensuring long-term durability.
\item \textbf{Materials Mechanics}: Investigates mechanical properties such as strength, ductility, and toughness, crucial for structural performance.
\item \textbf{Metallurgy}: Examines phase formation, phase diagrams, and microstructure control, offering insights into alloy stability and processing.
\item \textbf{Solid-State Physics}: Explores atomic-scale interactions, electronic structure, and thermal behavior, which influence phase stability and material performance.
\item \textbf{Materials Science}: Integrates insights from the other domains, focusing on the relationships between composition, structure, properties, and performance to optimize alloy design strategies.
\end{itemize}

The evidence collected from the LLM for each domain is mapped to one of four outcomes: \texttt{No knowledge}, \texttt{High}, \texttt{Medium}, or \texttt{Low}. These outcomes are translated into a corresponding mass function $m_{\mathrm{LLMs}}^{C_t,C_v}$, as shown in Table~\ref{tab:gpt-outcomes}. If the LLM indicates \texttt{No knowledge}, all mass is assigned to the set $\{\mathrm{similar}, \mathrm{dissimilar}\}$. If the LLM provides a specific substitutability rating (\texttt{High}, \texttt{Medium}, or \texttt{Low}), a portion of the mass is allocated to either $\{\mathrm{similar}\}$ or $\{\mathrm{dissimilar}\}$, with the remaining mass assigned to the $\Omega_{sim}$.

\subsection{Combining Evidence from Multiple Sources}
\label{subsec:combining-multiple-sources}

To integrate substitutability evidence collected from multiple sources, we employ Dempster's rule of combination with a \emph{reliability-aware discounting} step~\cite{Shafer1976,SMETS1993}. This step accounts for the reliability of each source with respect to our considered research target and adjusts the influence of each source's evidence, ensuring that more reliable sources contribute more to the final decision.

For each source $S$, we compute a dataset-specific discount factor:
\[
\gamma_{S} = \mathrm{disc}\!\bigl(m_S^{C_t,C_v}, \mathcal{D}\bigr) \;\in\; [0,1],
\]
where $\mathrm{disc}(.)$ quantifies how well the evidence about substitutability, collected from source \(S\), generalizes to property of alloys in \(\mathcal{D}\).  The reliability is evaluated using the macro-averaged F1 score in a $10$-fold cross-validation setup. For example, if a source \(S\) has historically shown accurate predictions on alloys that resemble those in \(\mathcal{D}\), we would set \(\gamma_S\) closer to \(1\). On the other hand, if \(S\) performs poorly or unpredictably for alloys in \(\mathcal{D}\), \(\gamma_S\) should be lowered.

The original mass function $m_S^{C_t,C_v}$ for source $S$, as defined in Sections~\ref{subsec:md-evidence} and~\ref{subsec:llm-evidence}, is thus adjusted to $^{\gamma_S}m_{S}^{C_t,C_v}$, as follows.
\begin{align}
^{\gamma_S}m_{S}^{C_t,C_v}(\{\mathrm{similar}\}) 
&= \gamma_S \, m_S^{C_t,C_v}\bigl(\{\mathrm{similar}\}\bigr),\\
^{\gamma_S}m_{S}^{C_t,C_v}(\{\mathrm{dissimilar}\}) 
&= \gamma_S \, m_S^{C_t,C_v}\bigl(\{\mathrm{dissimilar}\}\bigr),\\
^{\gamma_S}m_{S}^{C_t,C_v}(\Omega_{sim}) 
&= 1 - \gamma_S + \gamma_S \, m_S^{C_t,C_v}\bigl(\Omega_{sim}\bigr).
\end{align}

This redistributes mass from definitive conclusions \(\{\mathrm{similar}\}\) and \(\{\mathrm{dissimilar}\}\) to the ambiguous set \(\{\mathrm{similar}, \mathrm{dissimilar}\}\), encoding uncertainty for less reliable sources. As a result, when all mass functions are subsequently combined via Dempster's rule, a less trustworthy source exerts a weaker influence on the final outcome.

Suppose we have $p$ sources $\{S_1,S_2,\dots,S_p\}$. We combine pieces of evidence collected from them via the Dempster's rule of combination:
\begin{equation}
m^{C_t,C_v}(\omega)
\;=\;
\Bigl( ~^{\gamma_{S_1}}m_{S_1}^{C_t,C_v} \oplus ~^{\gamma_{S_2}}m_{S_2}^{C_t,C_v} \oplus \dots \oplus ~^{\gamma_{S_p}}m_{S_p}^{C_t,C_v}\Bigr)(\omega),
\end{equation}
where $\omega$ are nonempty subsets of $\Omega_{sim}$. 
The rule iteratively merges evidence while normalizing conflicts (e.g., empty-set intersections from contradictory sources). This preserves diversity in insights--from data-driven correlations to LLM-derived domain knowledge--while mitigating the impact of low-reliability sources.

Similar analyses are performed for all pairs of element combinations of interest to obtain a symmetric matrix $M$ consisting of all the similarities between them ($M[t,v]=M[v,t]=m^{C_t,C_v}(\{\mathrm{similar}\})$).

\begin{table*}[t]
\small
\centering
  \caption{\label{tab:dataset}Summary of alloy datasets used in evaluation experiments. No.\ alloys: Total number of alloys in each dataset. No.\ positive label: Number of alloys estimated to form HEA in the datasets $\mathcal{D}_{\text{0.9}T{m}}$ and $\mathcal{D}_{\text{1350K}}$, number of alloys with non-zero magnetization in $\mathcal{D}_{\text{Mag}}$, and number of alloys with non-zero Curie temperature in $\mathcal{D}_{T_C}$. The percentage values in parentheses indicate the proportion of positive labels in each dataset.}
  \label{tab:datasets}
  \begin{tabular*}{0.8\textwidth}{@{\extracolsep{\fill}}cccccc}
    \hline
    \textbf{Dataset} & \textbf{No.\ alloys} & \textbf{Physical properties} & \textbf{Positive label} & \textbf{No.\ positive label} \\
    \hline
    $\mathcal{D}_{\text{0.9}T_{m}}$~\cite{chen2023} & 14,950 quaternary alloys & Stability & HEA & 4,218 (28\%) \\
    $\mathcal{D}_{\text{1350K}}$~\cite{chen2023} & 14,950 quaternary alloys & Stability & HEA & 1,402 (9\%) \\
    $\mathcal{D}_{\text{Mag}}$~\cite{Ha2023} & 5,968 quaternary alloys & Magnetization ($T$) & Magnetic & 2,428 (41\%) \\
    $\mathcal{D}_{T_C}$~\cite{Ha2023} & 5,968 quaternary alloys & Curie temperature ($K$) & Nonzero Curie Temperature& 2,355 (39\%) \\
    \hline
\end{tabular*}
\end{table*}

\subsection{Evaluating Hypothetical Candidates by Analogy-Based Inference}

To predict whether a \emph{new} alloy $A_{new}$ is likely to form an HEA, we apply a substitution-based inference using the similarity matrix $M$. The process begins with a known alloy $A_k$ labeled $y_{A_k}$ and identifies the subset $C_t \subset A_k$ that, when replaced by $C_v$, generates the alloy $A_{new}$ (Fig.~\ref{fig:hybrid_framework} c). If $C_t$ and $C_v$ are substitutable, $y_{A_{new}}$ is more likely to match $y_{A_k}$; if they are dissimilar, $y_{A_{new}}$ may differ.

We formalize this inference using a frame of discernment~\cite{Shafer1976} $\Omega_{HEA} = \{\mathrm{HEA},\, \neg \mathrm{HEA}\}$
and define a mass function $m^{A_{new}}_{A_k, C_t \leftarrow C_v}$ to model evidence collected from $A_k$ and the substitution of $C_t,$ for $C_v$, denoted as $C_t \leftarrow C_v$. This mass function allocates belief to $\{ \mathrm{HEA} \}$, $\{ \neg \mathrm{HEA} \}$, or $\{ \mathrm{HEA},\, \neg \mathrm{HEA}\}$ according to the similarity $M[t,v]$ and the label of $A_k$:

\begin{equation}
m^{A_{new}}_{A_k, C_t \leftarrow C_v}\bigl(\{\mathrm{HEA}\}\bigr) =
  \begin{cases}
   M[t,v], & \text{if } y_{A_k} = \mathrm{HEA},\\
    0,      & \text{otherwise},
  \end{cases}
\end{equation}

\begin{equation}
m^{A_{new}}_{A_k, C_t \leftarrow C_v}\bigl(\{\neg \mathrm{HEA}\}\bigr) =
  \begin{cases}
    M[t,v], & \text{if } y_{A_k} = \neg \mathrm{HEA},\\
    0,       & \text{otherwise},
  \end{cases}
\end{equation}

\begin{equation}
m^{A_{new}}_{A_k, C_t \leftarrow C_v}\bigl(\Omega_{HEA}\bigr)
= 1 - M[t,v].
\end{equation}

Note that the masses assigned to $\{HEA\}$ and $\{\neg{HEA}\}$ reflect the levels of confidence whereby $A_k$ and the substitution of $C_v$ for $C_t$ support the probabilities that $A_{new}$ is or is not an HEA, respectively. The mass assigned to subset $\{HEA$, $\neg{HEA}\}$, expresses the probability that the evidence provide no information about the property of $A_{new}$. The sum of the probability masses assigned to all three nonempty subsets of $\Omega_{HEA}$ is 1.

Assume that we can collect multiple pieces of evidence, each derived from a different pair of host alloy $A_{host}$ and substitution pair $C_t \leftarrow C_v$, for a new alloy candidate $A_{new}$. These pieces of evidence can be combined using Dempster's rule of combination to produce a final mass function $m^{A_{new}}$. This function integrates all available analogies and resolves potential contradictions between the sources. The resulting combined evidence supports informed decision-making regarding whether to proceed with further costly experiments to validate the high-entropy alloy (HEA) formation ability of $A_{new}$.

\section{Results}

In this section, we present the design and outcomes of our experiments to evaluate both the \emph{predictive capability} and the \emph{interpretability} of our proposed method. Comparisons with alternative approaches (i.e., single-source evidential methods and other data-driven classifiers) are also provided. 

\subsection{Experimental Setting}
\label{subsec:exp-setting}

\begin{figure*}[t]
\centering
\includegraphics[width=\textwidth]{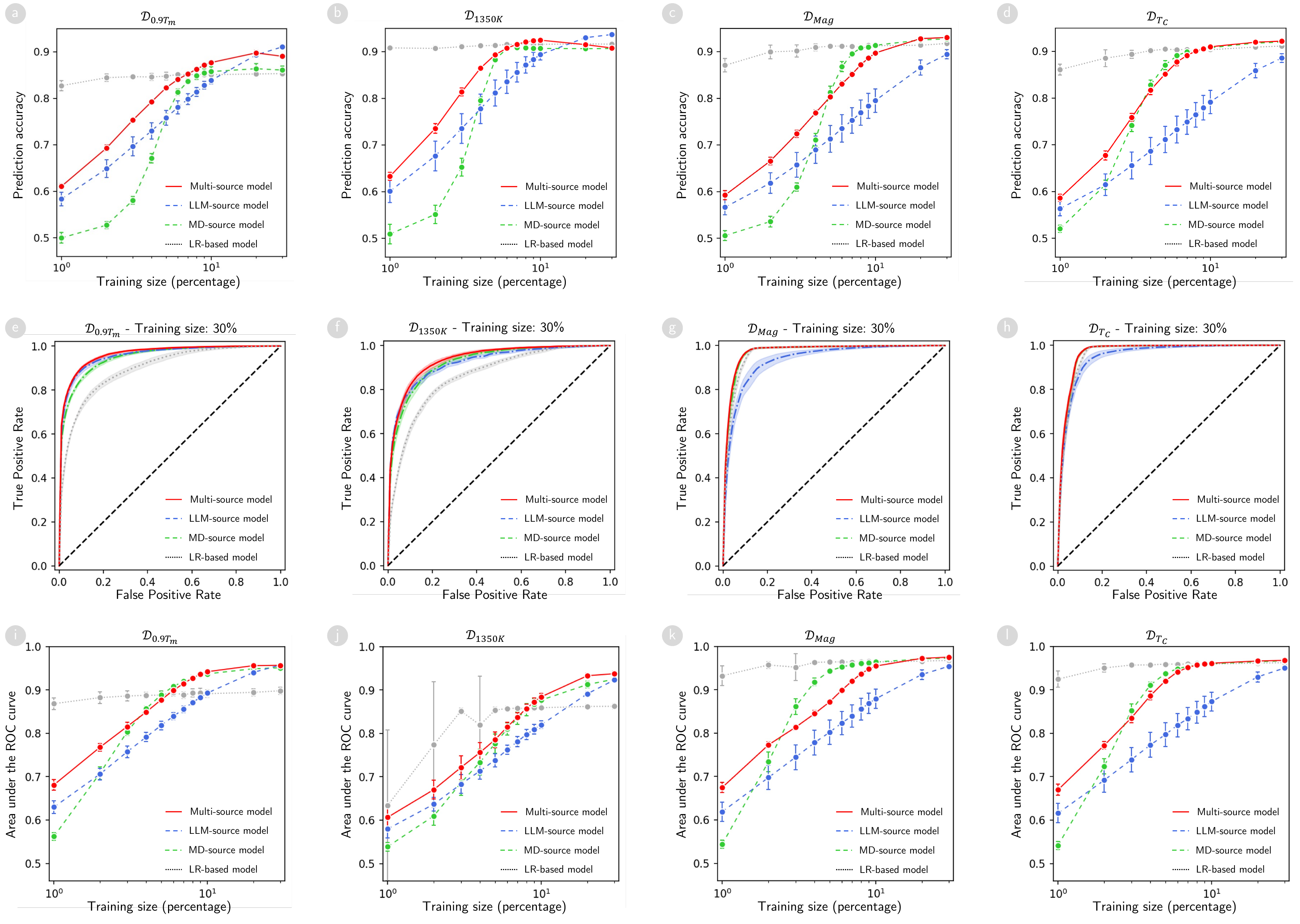}
\caption{\label{fig.exploitation_pa} Evaluation of predictive capability at varying training-set sizes.  
(a–d) Classification accuracy of the multi-source, single-source, andLR-based models on four quaternary-alloy datasets $\mathcal{D}_{0.9 T_m}$, $\mathcal{D}_{1350K}$, $\mathcal{D}_{\mathrm{Mag}}$, and $\mathcal{D}_{T_C}$.  (e–h) Receiver Operating Characteristic (ROC) curves for the same models at a 30\% training-set size on these datasets.  (i–l) The area under the ROC curves (AUC) for each model across a range of training-set sizes, providing an overall measure of discriminative performance.  In all subplots, the red lines indicate the multi-source model (using both MD and LLM sources), the green and blue lines denote single-source models (using either MD or LLM sources), and the gray lines represent the LR-based model.}
\end{figure*}

With the methodological foundation established, we now apply our proposed method to assess its predictive capabilities across various quaternary alloy datasets, evaluating both accuracy and interpretability. We evaluate our method on four computational datasets of quaternary alloys:
\begin{itemize}
  \item $\mathcal{D}_{0.9Tm}$ and $\mathcal{D}_{1350K}$: These datasets include \emph{all possible quaternary} alloys generated from the 26 elements in $\mathcal{E}_1$. The elemental set $\mathcal{E}_1$ comprises Fe, Co, Ir, Cu, Ni, Pt, Pd, Rh, Au, Ag, Ru, Os, Si, As, Al, Re, Mn, Ta, Ti, W, Mo, Cr, V, Hf, Nb, and Zr. Stability (i.e., whether an alloy forms a high-entropy alloy phase) is predicted based on methods proposed by Chen \emph{et al.}~\cite{chen2023} at two temperatures: $0.9\,T_m$ (approximately $90\%$ of the melting temperature $T_m$ of the alloy) and $1350\,( K)$. These predictions leverage a high-throughput computational workflow utilizing a regular-solution model~\cite{Takeuchi2005,Takeuchi2010} with binary interaction parameters extracted from \textit{ab initio} density-functional theory (DFT) to compare Gibbs free energies between solid solutions and competing intermetallic phases.
  
  \item $\mathcal{D}_{Mag}$ and $\mathcal{D}_{T_C}$\cite{Ha2023}: These datasets contain 5,968 equiatomic quaternary high-entropy alloys, generated from the element set $\mathcal{E}_2$, and the corresponding calculated properties: magnetizations ($\mathcal{D}_{Mag}$) and Curie temperatures ($\mathcal{D}_{T_C}$) in the body-centered cubic (BCC) phase. The elemental set $\mathcal{E}_2$ includes 21 transition metals: Fe, Co, Ir, Cu, Ni, Pt, Pd, Rh, Au, Ag, Ru, Os, Tc, Re, Mn, Ta, W, Mo, Cr, V, and Nb. These datasets were extracted from an original pool of $147,630$ equiatomic quaternary high-entropy alloys, computed using the Korringa-Kohn-Rostoker coherent approximation method~\cite{Fukushima2022}.
\end{itemize}

This study investigates the potential benefits of integrating multi-source knowledge derived from LLMs with material datasets to enhance predictions in exploration scenarios where specific elements are absent from the training data. To systematically evaluate the proposed method's predictive capability, we perform two complementary experiments:
\begin{enumerate}
    \item \textbf{Cross-validation on quaternary alloys:} we randomly partition each quaternary-alloy dataset into training and test subsets. The training set size varies from $1\%$ to $30\%$ of the dataset, enabling us to examine how the predictive accuracy of the methods scales with limited training data. This experiment examines how the alignment of LLM-derived knowledge with material-specific relationships in the material datasets, especially in data-scarce conditions. The goal is to assess whether the LLM-based evidence helps mitigate the challenges of sparse data and enhances prediction performance as more data becomes available.
    \item \textbf{Extrapolation on quaternary alloys:} we simulate an extrapolation scenario by excluding alloys containing a specific element $e$ from the training dataset. Models are trained on the remaining non-$e$ alloys and tested on $e$-containing alloys, mimicking real-world alloy discovery where researchers introduce novel elements into existing compositions. This experiment evaluates the adaptability of LLM-based knowledge to material-specific patterns and the ability of the proposed method to generalize effectively to unseen elements. By excluding alloys containing the specific element $e$ during training, we mimic the real-world scenario where researchers attempt to predict properties of novel alloys incorporating previously unobserved elements, simulating the discovery process.
\end{enumerate}

To benchmark our \emph{multi-source method}, we compare its predictive performance against:
\begin{itemize}
    \item \textbf{Single-source methods:} These use only one source of evidence (e.g., material dataset alone or just one domain knowledge collected from Gpt-4o).
    \item \textbf{Traditional classification methods:} Logistic Regression~\cite{Michael2008}. For the logistic-regression methods, we employ a compositional descriptor to numerically encode each alloy. Further information about the descriptor is provided in \textcolor{blue}{Supplementary Section III}.
\end{itemize}

Hyperparameters for the evidential methods (e.g., $\alpha$ and $\beta$, modeling substitutability from MD and LLM sources) are tuned via grid search, and competing machine learning methods, such as logistic regression, are optimized using their respective hyper-parameters (as shown in \textcolor{blue}{Supplementary Section IV}). After evaluating the predictive performance in these experiments, we analyze the substitutability of element combinations measured from our multi-source methods to explore the underlying mechanisms governing physical properties of quaternary alloys in the datasets. 

From now on, we will refer to models that utilize the evidential method, which applies Dempster-Shafer theory to create predictions based on evidence related to substitution mechanisms, as follows: models trained on evidence from material datasets are called MD-source models, those that leverage evidence from large language models (LLMs) are termed LLM-source models, and models that integrate evidence from both sources are known as multi-source models. Furthermore, model that uses Logistic Regression (LR) is referred to as LR-based model.

\begin{table*}[t]
\centering
\caption{\label{tab:extrapolation-results}Prediction accuracy of different methods on quaternary-alloy datasets for the extrapolation experiments. For each dataset, alloys containing a specific element $e$ are excluded during training and used as the test set. Results are reported as mean accuracy across all elements $e$ in these datasets.}
\begin{tabular*}{\textwidth}{@{\extracolsep{\fill}}lcccc}
\hline
\textbf{Methods} &  $\mathcal{D}_{\text{0.9}T_{m}}$ &$\mathcal{D}_{\text{1350K}}$ & $\mathcal{D}_{\text{Mag}}$ & $\mathcal{D}_{T_C}$ \\
\hline
Multi-source model & $\bm{0.87 \pm 0.06}$ & $\bm{0.92 \pm 0.04}$ & $\bm{0.86 \pm 0.19}$ & $\bm{0.86 \pm 0.18}$ \\
LLM-source model & $0.86 \pm 0.09$ & $0.91 \pm 0.08$ & $0.81 \pm 0.21$ & $0.86 \pm 0.18$ \\
MD-source model &  $0.50 \pm 0.03$ & $0.47 \pm 0.04$ & $0.48 \pm 0.07$ & $0.50 \pm 0.10$ \\
LR-based model & $0.84 \pm 0.06$ & $0.91 \pm 0.04$ & $0.67 \pm 0.15$ & $0.68 \pm 0.13$ \\
\hline
\end{tabular*}
\end{table*}

\begin{table*}[t]
\centering
\caption{\label{tab:auc-results}Areas under the Receiver Operating Characteristic (ROC) curves of different methods on the four quaternary-alloy datasets for the extrapolation experiments. For each dataset, alloys containing a specific element $e$ are excluded during training and used as the test set. Results are reported as mean areas across all elements $e$ in these datasets.}
\begin{tabular*}{\textwidth}{@{\extracolsep{\fill}}lcccc}
\hline
\textbf{Methods} &  $\mathcal{D}_{\text{0.9}T_{m}}$ &$\mathcal{D}_{\text{1350K}}$ & $\mathcal{D}_{\text{Mag}}$ & $\mathcal{D}_{T_C}$ \\
\hline
Multi-source model & $\bm{0.93 \pm 0.06}$ & $\bm{0.92 \pm 0.08}$ & $\bm{0.95 \pm 0.06}$ & $\bm{0.94 \pm 0.07}$ \\
LLM-source model  & $0.91 \pm 0.11$ & $0.90 \pm 0.12$ & $0.95 \pm 0.06$ & $0.94 \pm 0.07$ \\
MD-source model &  $0.50 \pm 0.00$ & $0.50 \pm 0.00$ & $0.50 \pm 0.00$ & $0.50 \pm 0.00$ \\
LR-based model & $0.84 \pm 0.12$ & $0.83 \pm 0.11$ & $0.84 \pm 0.06$ & $0.84 \pm 0.06$ \\
\hline
\end{tabular*}
\end{table*}

\subsection{Evaluation of Predictive Capability by Cross-Validation}
\label{subsec:cv}

\begin{figure*}[t]
\centering
\includegraphics[width=\textwidth]{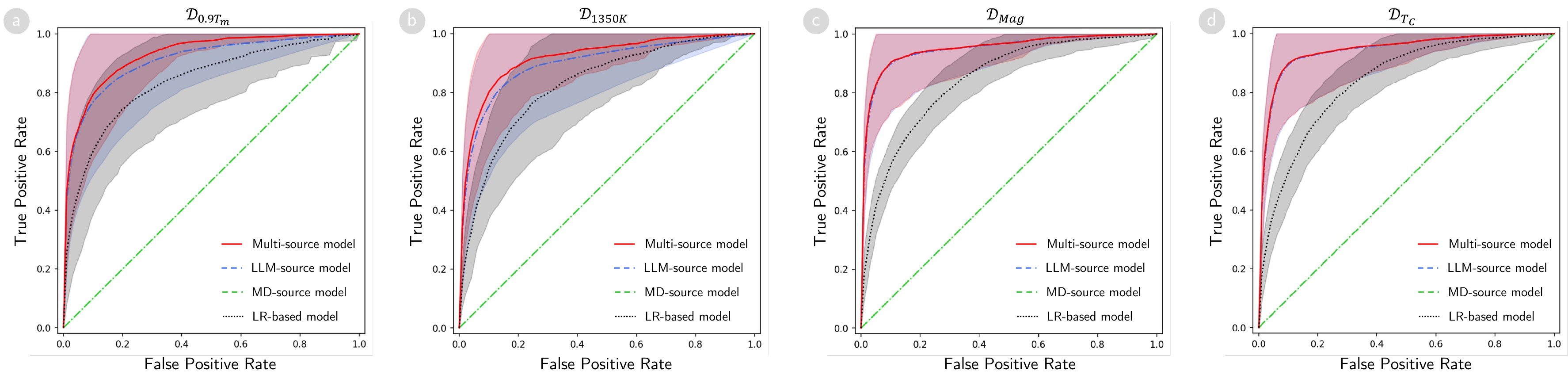}
\caption{\label{fig.exploration_pa} Predictive capability evaluation through extrapolation on four quaternary-alloy datasets. For each dataset, alloys containing a specific element $e$ are excluded from training and used as the test set. (a-d) The area under the ROC curves (AUC) is shown for each model on the respective test sets in the extrapolation experiments. In all subplots, the red lines indicate the multi-source model (integrating both MD and LLM sources), the green and blue lines denote single-source models (using either MD or LLM sources), and the gray lines represent the LR-based model.}
\end{figure*}

In this experiment, we vary the training set size from 1\% to 30\% of each quaternary-alloy dataset (stepping in 1\% increments up to 10\%, then 20\% and 30\%), enabling us to examine how methods cope with data scarcity versus moderate availability. Figure \ref{fig.exploitation_pa} shows that at smaller training sizes (approximately 1\% -- 10\%), logistic regression achieves the highest overall accuracy, surpassing evidential models, which explicitly model the element substitutability to predict physical properties of the alloys. Among the evidential models, single LLM-source models initially outperform MD-source models, likely because the LLM's domain-specific insights help mitigate data limitations. However, multi-sources model remain competitive, and can sometimes yield the top accuracy among evidential models even under limited data. As the training size exceeds 10\%, MD-source models exhibit superior performance on magnetization and Curie temperature datasets, while reaching accuracy levels comparable to LLM-source models on alloy stability datasets. By contrast, logistic regression's accuracy plateaus and is eventually eclipsed by evidential models. These trends highlight the value of incorporating knowledge from either an LLM source, an MD source, or both to improve quaternary-alloy property predictions.

Although the prediction accuracy provides a convenient single-metric overview, it relies on a fixed classification threshold--commonly 0.5--that may be suboptimal in imbalanced datasets, where high-entropy alloys (positive class) are relatively rare. Under such conditions, logistic regression can appear particularly strong at very small training sizes if it effectively defaults to predicting the dominant (Non-HEA) class, thus inflating accuracy. Such an approach cannot address scenarios where different types of misclassifications (e.g., false positives versus false negatives) incur different costs. To more fully capture these trade-offs under dynamic thresholds, we analyze the Receiver Operating Characteristic (ROC) curves of the methods on four datasets (Fig.~\ref{fig.exploitation_pa} e-l), which show how each method's true positive rate (TPR) and false positive rate (FPR) vary across all possible decision boundaries. 

Figure~\ref{fig.exploitation_pa} e-h presents the ROC curves for the multi-source models, LLM-source models, MD-source models, and LR-based models at a 30\% training size across the four datasets.  Overall, the multi-sources models and the MD-source models exhibit similar ROC performance and outperform the other models. The LLM-source models achieve results comparable to the best ones on the and stability datasets $\mathcal{D}_{\text{0.9}T_{m}}$ and $\mathcal{D}_{\text{1350K}}$, but lag behind the MD-source models on the datasets related to magnetization and Curie temperatures $\mathcal{D}_{\text{Mag}}$ and $\mathcal{D}_{T_C}$. This finding suggests that the knowledge collected from the five considered research domains may not fully align with the magnetic and thermal properties reflected in those datasets. In contrast, logistic regression exhibits the lowest performance on the stability datasets. On the magnetization and Curie temperature datasets, however, its performance is marginally better than that of the lowest-performing models, the LLM-source models. 

To further examine each model’s ROC performance at different amounts of training data, we analyze the AUC distribution from 1\% up to 30\% training size (Fig.~\ref{fig.exploitation_pa} i-k). When the training set is very small, models drawing upon LLM-based knowledge often enjoy an early advantage, presumably because domain insights can compensate for limited alloy observations. However, as more data accumulates, MD-source models typically surpass these LLM-source models, suggesting that direct data-driven cues from the quaternary-alloy sets become increasingly decisive. In contrast, multi-source models—which combine both material-dataset evidence and LLM-derived insights—maintain robust performance across all training-set sizes, reflecting the flexibility gained by merging domain-based substitutability perspectives with empirical data. Multi-sources models leverage complementary evidence, enabling them to balance true positive and false positive rates more effectively. On the stability datasets $\mathcal{D}_{\text{0.9}T_{m}}$ and $\mathcal{D}_{\text{1350K}}$, MD-source and multi-source models attain comparable AUC early on and remain highly competitive as the training size grows. Meanwhile, in magnetization and Curie-temperature datasets, MD-source models briefly surpass multi-source models at moderate training sizes (roughly 6–20\%), but this gap diminishes at the largest training sizes. 

In conclusion, models utilizing knowledge derived from LLM excel in data-scarce scenarios by utilizing domain-specific insights to mitigate sparse data limitations. As data availability increases, MD-source models surpass LLM-source models, particularly where MD evidence provide sufficiently for the data-driven approach learn models aligns more closely with these underlying processes. Multi-source models, which integrate both LLM and MD insights, demonstrate strong and consistent performance across various training sizes but occasionally underperform at intermediate data scales, indicating the need for calibrated integration strategies. 

\subsection{Evaluation of Predictive Capability by Extrapolation}
\label{subsec:extrapolation}

Having assessed our models via cross-validation (Section~\ref{subsec:cv}), we now examine how well they \emph{extrapolate} to quaternary alloys containing an element $e$ that was excluded during training. In contrast to the cross-validation experiments, we do not vary the training set size here. Instead, for each element $e$, we remove all $e$-containing alloys from the dataset and train each model on the remaining alloys, which do not contain the element $e$. We then evaluate performance on the $e$-containing alloys. This procedure tests whether the learned models can predict properties for compositions involving an element never observed in their training data.

Table~\ref{tab:extrapolation-results} shows that the evidential models relying solely on the material-dataset-based evidence (MD-based evidence) achieve relatively poor accuracies across the four datasets, ranging from $0.47$ to $0.56$.  In contrast, the models that incorporate knowledge from either a single LLM source or from multiple evidence sources attain significantly higher accuracies on each dataset: for instance, on $\mathcal{D}_{\text{0.9}T_{m}}$ they reach $0.86$ and $0.87$, respectively, compared to just $0.50$ for the MD-source models. This improvement is particularly noteworthy on $\mathcal{D}_{\text{Mag}}$ and $\mathcal{D}_{T_C}$, where the single-LLM and multi-source evidential methods surpass both the LR-based models and the MD-source models by a large margin. Notably, the multi-source models achieve the best performance overall on all the considered datasets, consistently outperforming the LLM-source models by a modest but persistent gap.

Figure~\ref{fig.exploration_pa} illustrates the ROC curves for the multi-sources and LLM-source models consistently lie closer to the top-left corner, indicating higher true positive rates at comparable false positive rates over all datasets. In contrast, the curves for MD-source models lie on the diagonal, signifying near-random discrimination, while logistic regression yields moderate performance between these extremes. To quantify these visual differences, Table~\ref{tab:auc-results} lists the area under the ROC curve (AUC) for each dataset, revealing that multi-sources models achieve the highest scores (0.92--0.95) across the four datasets, followed closely by LLM-source models (0.90--0.95). By comparison, LR-based models peaks around 0.84, whereas MD-source models hovers at 0.50. Overall, these results confirm that leveraging multiple or LLM-based evidence sources substantially improves discriminative power under the extrapolation scenario.

\begin{figure*}[t]
\centering
\includegraphics[width=\textwidth]{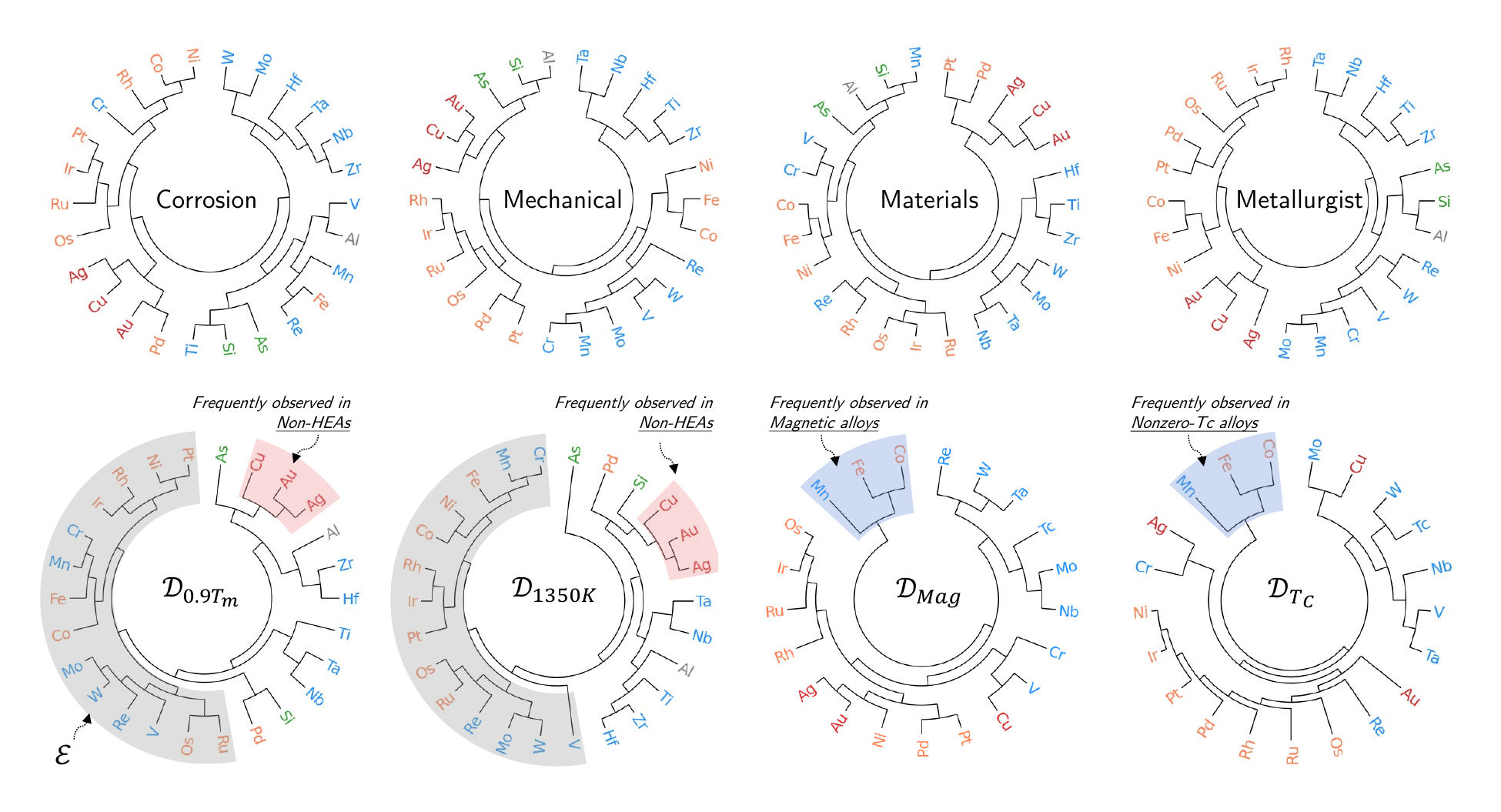}
\caption{\label{fig.substitutability_matrix} Circular hierarchical clustering (HAC) of elements based on substitutability between elements.The circular dendrogram displays the hierarchical clustering of all constituent elements, constructed using hierarchical agglomerative clustering (HAC) with the "complete" linkage criterion. The substitutability information is derived from both alloy datasets and LLM-based knowledge. Blue labels represent early transition metals, orange labels indicate late transition metals, and red labels denote coinage metals, including copper (Cu), silver (Ag), and gold (Au).}
\end{figure*}

\begin{figure*}[t]
\centering
\includegraphics[width=\textwidth]{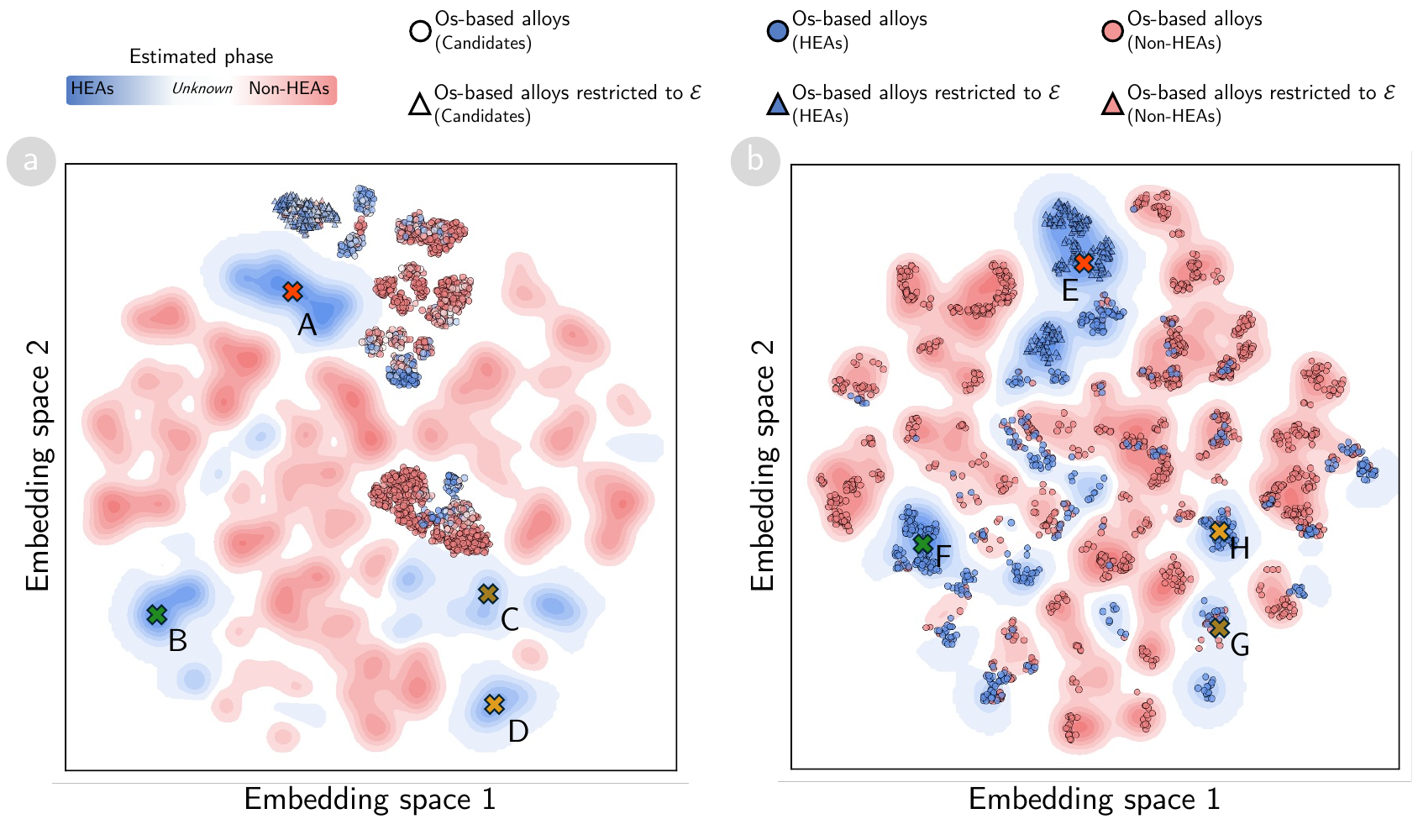}
\caption{\label{fig.alloys_map} Alloy maps from the dataset $\mathcal{D}_{0.9T_m}$ visualizing extrapolation simulations for Os-based alloys.Triangular markers indicate Os-based alloys with elements restricted to $\mathcal{E}$, while circular markers represent Os-containing alloys with elements potentially outside $\mathcal{E}$. (a) Map constructed using similarity information excluding Os-based alloys. Marker colors indicate predicted phase behavior: blue for higher belief in HEAs formation, red for Non-HEAs formation, and white for regions of high uncertainty.(b) Updated map after integrating Os-based alloys into the dataset. Blue points denote Os-based alloys predicted to form HEAs, while red points indicate Non-HEAs formation. The visualization highlights structural changes and reorganization in alloy groups following the inclusion of Os-based alloys.}
\end{figure*}

\subsection{Learning about the Substitutability between Elements}
\label{subsec:substitutability}

The previous experiments demonstrate that integrating multi-source knowledge enables robust extrapolation in predicting HEA formation, even with novel elements. In this section, we perform further analysis of the substitutability between elements collected from both MD and LLM sources to provide a detailed explanation of the high performance observed. Figure \ref{fig.substitutability_matrix} presents a hierarchically clustered structure of constituent elements, constructed based on substitutability information. This structure is generated by applying hierarchical agglomerative clustering (HAC) with the complete linkage criterion. Elements within the same cluster exhibit similar substitutability patterns.

The structure derived from domain knowledge obtained through LLMs reveals a strong consensus across domains regarding the coinage metals—copper (Cu), silver (Ag), and gold (Au)—which consistently exhibit similar substitutability behavior. Other transition metals also form subclusters, though the exact clustering can vary across different domain perspectives. However, a key and recurring pattern is the clear distinction between early transition metals (blue-labeled) and late transition metals (orange-labeled). Furthermore, aluminum (Al), silicon (Si), and arsenic (As) cluster together based on substitutability information, indicating shared behavior. Transition metals from groups 4 and 5, including tantalum (Ta), niobium (Nb), hafnium (Hf), titanium (Ti), and zirconium (Zr), tend to form another distinct cluster, suggesting stronger substitutability relationships within this group. 

Figure \ref{fig.substitutability_matrix} presents the hierarchically clustered structure of constituent elements, constructed from substitutability information obtained through LLMs across four domains: Corrosion, Mechanical Properties, Materials Science, and Metallurgy. Notably, the structure derived from Solid-State Physics is similar to that of Metallurgy. Detailed results from Solid-State Physics are provided in \textcolor{blue}{Supplementary Figure 3}.

We observe similar patterns from the substitutability data collected in the datasets $\mathcal{D}_{\text{0.9}T{m}}$ and $\mathcal{D}_{\text{1350K}}$. The coinage metals again form a cohesive group, but they display a negative synthesized effect on phase stability: most alloys containing Cu, Ag, or Au do not form HEAs under both temperature conditions, as shown in \textcolor{blue}{Supplementary Figure 1}. Additionally, the group of 14 transition metals, denoted as $\mathcal{E}$, demonstrates consistent substitutability behavior. Further investigation into the synthesized effects of this group will be discussed more details in the following section (Section \ref{subsec:insights}).

Despite general consistency with domain knowledge, certain elements display divergent substitutability behavior. Aluminum (Al) and silicon (Si) exhibit patterns similar to those of transition metals from groups 4 and 5, as well as palladium (Pd), under both temperature conditions. In contrast, arsenic (As) behaves distinctly, separating from both metalloid and transition metal clusters. This deviation impacts predictive performance during extrapolation experiments where alloys containing one of these elements are excluded from the training set, as illustrated in \textcolor{blue}{Supplementary Figure 2}.

Substitutability data from the magnetization and Curie temperature datasets, $\mathcal{D}_{\text{Mag}}$ and $\mathcal{D}_{T_C}$, reveal two dominant clusters based on early and late transition metals, consistent with other sources. However, manganese (Mn), iron (Fe), and cobalt (Co) form a separate subcluster, showing positive effects on magnetic properties such as magnetization and Curie temperature (\textcolor{blue}{Supplementary Figure 1}). This subcluster diverges from LLM-based substitutability data, impacting the performance of our hybrid framework when exploring these metals. \textcolor{blue}{Supplementary Figure 2} shows that predictive performance for Mn, Fe, and Co is significantly lower than for other elements in the magnetization and Curie temperature datasets.

In summary, the analysis reveals distinct groups of elements exhibiting consistent behavior across stability and key physical properties, such as magnetization and Curie temperature. These findings emphasize that while integrating domain knowledge enhances exploration scenarios within the hybrid framework, alignment with underlying physical mechanisms is crucial to fully support robust predictions.

\subsection{Towards Insights into the HEA Formation Mechanism}
\label{subsec:insights}

Alloy design decisions carry significant costs and risks, requiring decision support systems to provide both high predictive accuracy and interpretable insights. To address this, we propose a t-SNE (t-distributed Stochastic Neighbor Embedding) visualization that organizes alloys by shared compositions and phase behavior. The visualization employs a hybrid distance matrix that integrates (1) dissimilarity scores from empirical and knowledge-based frameworks and (2) Jaccard distances, which capture compositional overlap. This approach reveals chemical relationships and mechanisms underlying phase stability, aiding hypothesis generation and validation. Detailed information on the hybrid distance is provided in \textcolor{blue}{Supplementary Section V}. This section also explores the role of the $\mathcal{E}$ set of 14 transition metals, which exhibit strong substitutability and are critical to HEA stability. We analyze the synthesized effect of $\mathcal{E}$—specifically, how its elements contribute to or interact with other components to influence phase stability—through visual analysis.

\begin{figure*}[t]
\centering
\includegraphics[width=\textwidth]{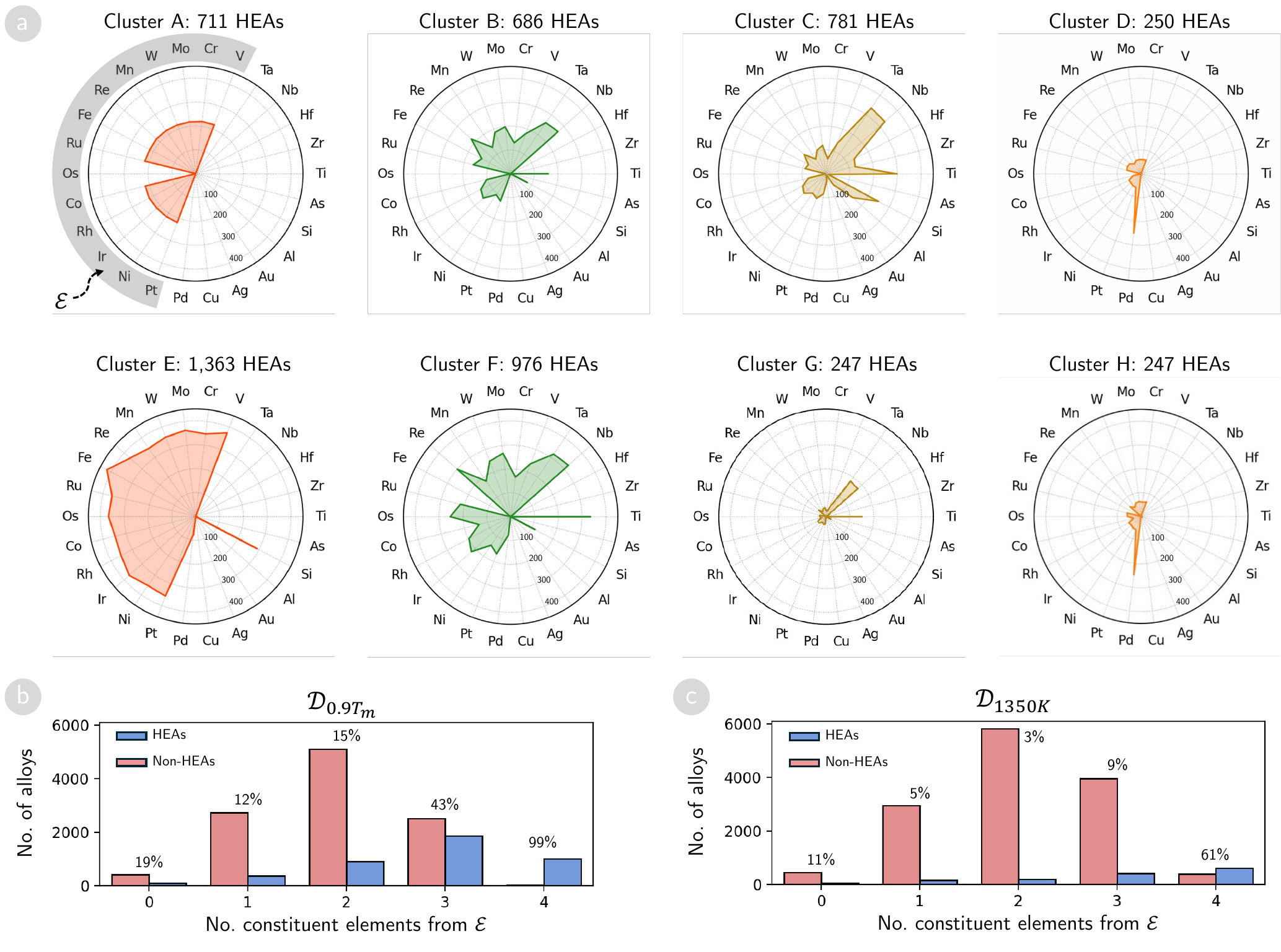}
\caption{\label{fig.cluster_profile} Summary of HEA groups and analysis on impact of constituent elements from element set $\mathcal{E}$. (a) The radar charts depict the distribution of constituent elements in high-entropy alloys (HEAs) across groups formed by quaternary alloys. Groups A, B, C, and D represent HEAs without osmium (Os), while groups E, F, G, and H include compositions formed after integrating Os-based alloys. (b-c) The bar plots illustrate the number of HEAs and Non-HEAs, showing how stability varies with the number of constituent elements from $\mathcal{E}$ on the two datasets $\mathcal{D}_{0.9T_m}$ (b) and $\mathcal{D}_{\text{1350K}}$ (c). The percentage values indicate the rates of HEAs under each configuration.}
\end{figure*}

Figures \ref{fig.alloys_map} a and b illustrate the maps of quaternary alloys from the dataset $\mathcal{D}_{0.9T_m}$, visualizing extrapolation simulations for Os-based alloys, which were not included in the training data. Figure \ref{fig.alloys_map} a depicts the map constructed using similarity information from $\mathcal{D}_{0.9T_m}$ without Os-based alloys, while fig. \ref{fig.alloys_map} b incorporates updated similarity information after integrating the labels of Os-based alloys into the dataset.

Four distinct regions with blue backgrounds are evident in fig. \ref{fig.alloys_map} a,, indicating areas dominated by observed high-entropy alloys (HEAs). These HEA-forming alloys converge into four groups, labeled A, B, C, and D. Figure \ref{fig.cluster_profile} a provides a detailed summary of the constituent elements for each group. Group A stands out with 711 HEAs, all formed from a set of 13 critical transition metals, which is a subset of $\mathcal{E}$. The full set $\mathcal{E}$, including Os, is identified in Section \ref{subsec:substitutability}. This subset accounts for 99\% of all possible quaternary combinations derived from these elements. The near-complete coverage of combinatorial possibilities underscores the pivotal role these 13 transition metals play in promoting HEA formation. Their consistent presence across successful alloy compositions positions them as a core foundation for phase stability and robust alloy design.

In contrast,groups B and C are characterized by alloys with niobium (Nb), tantalum (Ta), titanium (Ti), and silicon (Si), combined with the elements from the subset of $\mathcal{E}$. Group D, by contrast, comprises 250 palladium (Pd)-based alloys, where Pd is alloyed with the 13 elements in the subset. The group D represent 87\% of the 286 possible quaternary alloy configurations, underscoring Pd’s versatility in this compositional space. Notably, a significantly reduced cobalt (Co) content within group D suggests diminished synergistic interactions between Co and Pd when integrated with other elements in the subset of $\mathcal{E}$.

In Figure \ref{fig.alloys_map} a, circular and triangular markers highlight osmium (Os)-containing alloys absent from the training data. Marker colors reflect uncertainty levels derived from domain knowledge distilled via the GPT-4o model. Domain knowledge indicates that Os shares strong atomic and chemical similarities with remaining metals in $\mathcal{E}$ \textcolor{blue}{(Supplementary Figure 3)}, rationalizing the high confidence (blue triangular markers above group A) in HEA formation for Os-based alloys restricted to elements in $\mathcal{E}$. This hypothesis is further corroborated by an 88\% prediction accuracy for Os-based alloys \textcolor{blue}{(Supplementary Table 1)}, demonstrating strong alignment between Os and the remaining element in $\mathcal{E}$.

Figure \ref{fig.alloys_map} b presents the updated alloy map after integrating Os-based alloys. At first glance, four groups of HEAs are evident, labeled as groups E, F, G, and H. A summary of these groups is provided in Fig. \ref{fig.cluster_profile} a. Group E contains a similar proportion of elements to group A, with the addition of Os, Si, and part of Pd. Further analysis reveals that Group E forms through interactions between Os-based alloys restricted to $\mathcal{E}$ and HEAs in Group A, as shown in \textcolor{blue}{Supplementary Figure 4}. Additionally, group E also incorporates HEAs originating from groups C and D, including alloys from the Si-based and Pd-based alloys, respectively. In contrast, group F is derived from group B, now adding Os-based alloys, while group G and H are formed from the remaining HEAs in groups C and D, respectively. The emergence of group E highlights the strong similarity between Os and remaining elements in $\mathcal{E}$. Of the 1,001 quaternary alloys in the $\mathcal{E}$, 997 form HEAs, achieving a 99\% stability rate. 

Additional analyses are performed to investigate further the impact of the set $\mathcal{E}$ on the HEA formation mechanism. Figures \ref{fig.cluster_profile}b and \ref{fig.cluster_profile}c illustrate the number of HEAs and Non-HEAs as the number of constituent elements from $\mathcal{E}$ decreases. Nearly 98\% of all possible HEAs in the dataset contain at least one element from the set $\mathcal{E}$ of 14 transition metals, emphasizing the foundational role of these elements in stabilizing HEAs. When alloys are formed exclusively from elements in this set, 99\% form HEAs; however, the stability rate declines as fewer elements from $\mathcal{E}$ are included (Fig. \ref{fig.cluster_profile}b).  \textcolor{blue}{Supplementary Figure 5} provides further details, indicating that the stability rate of alloys formed by combining elements from $\mathcal{E}$ with one of remaining elements ranges from 0\% to 84\%. 

This strong stabilizing effect of the element set $\mathcal{E}$ persists even at elevated temperatures, as shown by the dataset at 1350 (K) in Fig. \ref{fig.cluster_profile}c. In this dataset, nearly 97\% of all possible HEAs still contain at least one element from the set, but the stability rate drops to 61\% when only the elements from $\mathcal{E}$ are used to form alloys.

In summary, the results emphasize that the identified set of 14 transition metals forms a critical foundation for phase stability in HEAs. The visualization demonstrates how both novel and known elements integrate into HEA, linking compositional effects to phase stability. This approach enhances interpretability and supports alloy design by providing actionable, knowledge-driven insights into phase formation tendencies.

\section*{Conclusions}

This study introduces a hybrid framework that integrates multi-source knowledge from large language models (LLMs) and material datasets (MD) to support decision-making in high-entropy alloy (HEA) discovery. The framework centers around the principle of elemental substitutability, providing an interpretable approach to exploring compositional spaces. By aggregating evidence from both empirical data and domain knowledge, the framework effectively addresses challenges related to data scarcity, uncertainty, and exploration.

Our results demonstrate that incorporating multi-source knowledge achieves strong performance across various interpolation scenarios. However, the framework’s strength is particularly evident under extrapolation conditions, where integrating diverse evidence sources significantly enhances predictive performance. Models utilizing LLM-derived knowledge excel in data-sparse situations, leveraging domain expertise to bridge information gaps. As training data grows, MD-based models eventually outperform LLM-based models due to their closer alignment with property-specific mechanisms. This observation emphasizes the need for dynamic strategies that balance contributions from different evidence sources based on dataset characteristics.

To further elucidate HEA formation mechanisms, we developed a visualization of compositional landscapes informed by substitutability data. This visualization reveals distinct clusters of alloys, illustrating how elemental combinations influence phase stability. Our findings highlight the pivotal role of a core set of 14 transition metals, denoted as $\mathcal{E}$, which consistently exhibit strong substitutability relationships. These elements form a robust foundation for stabilizing single-phase HEAs, with their stabilizing effect persisting across various temperature ranges. This insight underscores their critical importance in alloy design and optimization.

Despite these promising advances, challenges remain in optimizing the integration of multi-source evidence. In some cases, multi-source models underperform at intermediate training sizes, suggesting that evidence integration requires careful calibration. Future work should focus on adaptive frameworks that dynamically adjust evidence weighting based on real-time performance validation. Additionally, expanding the framework to include diverse material properties—such as mechanical strength or thermal stability—will further enhance its applicability.

Overall, this study demonstrates the potential of combining data-driven approaches with expert-derived knowledge to improve both prediction accuracy and interpretability in HEA discovery. By systematically managing uncertainties and encouraging exploration in data-scarce regions, the proposed framework advances the frontier of materials science, accelerating the discovery of innovative alloy compositions.

\section*{Author contributions}
\textbf{M.-Q.H.}: Conceived and designed the experiments, Performed the experiments, Analyzed the data, Contributed materials/analysis tools, Wrote the paper. 
\textbf{D.-K.L.}: Conceived and designed the experiments, Performed the experiments, Analyzed the data, Wrote the paper. 
\textbf{D.-A.D.}: Conceived and designed the experiments, Analyzed the data.
\textbf{T.-S.V.}: Performed the experiments, Analyzed the data.
\textbf{D.-N.N.}: Conceived and designed the experiments, Analyzed the data.
\textbf{V.-C.N.}: Performed the experiments, Contributed materials/analysis tools.
\textbf{H.K.}: Conceived and designed the experiments, Analyzed the data, Wrote the paper. 
\textbf{V.-N.H.}: Conceived and designed the experiments, Writing -- Review \& Editing. 
\textbf{H.-C.D.}: Conceived and designed the experiments, Performed the experiments, Analyzed the data, Contributed materials/analysis tools, Wrote the paper. 

\section*{Conflicts of interest}
The authors declare no competing interests.

\section*{Data availability}
Data available on request from the authors

\section*{Acknowledgements}
This work is supported by the JSPS KAKENHI Grants 23KJ1035.


\balance

\bibliography{main.bib} 

\providecommand*{\mcitethebibliography}{\thebibliography}
\csname @ifundefined\endcsname{endmcitethebibliography}
{\let\endmcitethebibliography\endthebibliography}{}
\begin{mcitethebibliography}{38}
\providecommand*{\natexlab}[1]{#1}
\providecommand*{\mciteSetBstSublistMode}[1]{}
\providecommand*{\mciteSetBstMaxWidthForm}[2]{}
\providecommand*{\mciteBstWouldAddEndPuncttrue}
  {\def\EndOfBibitem{\unskip.}}
\providecommand*{\mciteBstWouldAddEndPunctfalse}
  {\let\EndOfBibitem\relax}
\providecommand*{\mciteSetBstMidEndSepPunct}[3]{}
\providecommand*{\mciteSetBstSublistLabelBeginEnd}[3]{}
\providecommand*{\EndOfBibitem}{}
\mciteSetBstSublistMode{f}
\mciteSetBstMaxWidthForm{subitem}
{(\emph{\alph{mcitesubitemcount}})}
\mciteSetBstSublistLabelBeginEnd{\mcitemaxwidthsubitemform\space}
{\relax}{\relax}

\bibitem[Yeh \emph{et~al.}(2004)Yeh, Chen, Lin, Gan, Chin, Shun, Tsau, and
  Chang]{Yeh2004}
J.-W. Yeh, S.-K. Chen, S.-J. Lin, J.-Y. Gan, T.-S. Chin, T.-T. Shun, C.-H. Tsau
  and S.-Y. Chang, \emph{Advanced Engineering Materials}, 2004, \textbf{6},
  299--303\relax
\mciteBstWouldAddEndPuncttrue
\mciteSetBstMidEndSepPunct{\mcitedefaultmidpunct}
{\mcitedefaultendpunct}{\mcitedefaultseppunct}\relax
\EndOfBibitem
\bibitem[Cantor \emph{et~al.}(2004)Cantor, Chang, Knight, and
  Vincent]{CANTOR2004213}
B.~Cantor, I.~Chang, P.~Knight and A.~Vincent, \emph{Materials Science and
  Engineering: A}, 2004, \textbf{375-377}, 213 -- 218\relax
\mciteBstWouldAddEndPuncttrue
\mciteSetBstMidEndSepPunct{\mcitedefaultmidpunct}
{\mcitedefaultendpunct}{\mcitedefaultseppunct}\relax
\EndOfBibitem
\bibitem[Senkov \emph{et~al.}(2015)Senkov, Miller, Miracle, and
  Woodward]{Senkov2015}
O.~N. Senkov, J.~D. Miller, D.~B. Miracle and C.~Woodward, \emph{Nat. Commun.},
  2015, \textbf{6}, 6529\relax
\mciteBstWouldAddEndPuncttrue
\mciteSetBstMidEndSepPunct{\mcitedefaultmidpunct}
{\mcitedefaultendpunct}{\mcitedefaultseppunct}\relax
\EndOfBibitem
\bibitem[Rickman \emph{et~al.}(2019)Rickman, Chan, Harmer, Smeltzer, Marvel,
  Roy, and Balasubramanian]{Rickman2019}
J.~M. Rickman, H.~M. Chan, M.~P. Harmer, J.~A. Smeltzer, C.~J. Marvel, A.~Roy
  and G.~Balasubramanian, \emph{Nature Communications}, 2019, \textbf{10},
  2618\relax
\mciteBstWouldAddEndPuncttrue
\mciteSetBstMidEndSepPunct{\mcitedefaultmidpunct}
{\mcitedefaultendpunct}{\mcitedefaultseppunct}\relax
\EndOfBibitem
\bibitem[Tsai and Yeh(2014)]{Tsai2014}
M.-H. Tsai and J.-W. Yeh, \emph{Materials Research Letters}, 2014, \textbf{2},
  107--123\relax
\mciteBstWouldAddEndPuncttrue
\mciteSetBstMidEndSepPunct{\mcitedefaultmidpunct}
{\mcitedefaultendpunct}{\mcitedefaultseppunct}\relax
\EndOfBibitem
\bibitem[Deshmukh \emph{et~al.}(2024)Deshmukh, Wichrowski, Evangelou, Ghanekar,
  Deshpande, Kevrekidis, and Greeley]{Gaurav2024}
G.~Deshmukh, N.~J. Wichrowski, N.~Evangelou, P.~G. Ghanekar, S.~Deshpande,
  I.~G. Kevrekidis and J.~Greeley, \emph{npj Computational Materials}, 2024,
  \textbf{10}, 116\relax
\mciteBstWouldAddEndPuncttrue
\mciteSetBstMidEndSepPunct{\mcitedefaultmidpunct}
{\mcitedefaultendpunct}{\mcitedefaultseppunct}\relax
\EndOfBibitem
\bibitem[Ghorbani \emph{et~al.}(2024)Ghorbani, Boley, Nakashima, and
  Birbilis]{Ghorbani2024}
M.~Ghorbani, M.~Boley, P.~N.~H. Nakashima and N.~Birbilis, \emph{Scientific
  Reports}, 2024, \textbf{14}, 8299\relax
\mciteBstWouldAddEndPuncttrue
\mciteSetBstMidEndSepPunct{\mcitedefaultmidpunct}
{\mcitedefaultendpunct}{\mcitedefaultseppunct}\relax
\EndOfBibitem
\bibitem[Tsai(2016)]{Tsai2016}
M.-H. Tsai, \emph{Entropy}, 2016, \textbf{18}, 252\relax
\mciteBstWouldAddEndPuncttrue
\mciteSetBstMidEndSepPunct{\mcitedefaultmidpunct}
{\mcitedefaultendpunct}{\mcitedefaultseppunct}\relax
\EndOfBibitem
\bibitem[Tsai \emph{et~al.}(2019)Tsai, Tsai, Chang, and Huang]{Tsai2019}
M.-H. Tsai, R.-C. Tsai, T.~Chang and W.-F. Huang, \emph{Metals}, 2019,
  \textbf{9}, 247\relax
\mciteBstWouldAddEndPuncttrue
\mciteSetBstMidEndSepPunct{\mcitedefaultmidpunct}
{\mcitedefaultendpunct}{\mcitedefaultseppunct}\relax
\EndOfBibitem
\bibitem[Huang \emph{et~al.}(2019)Huang, Martin, and Zhuang]{HUANG2019225}
W.~Huang, P.~Martin and H.~L. Zhuang, \emph{Acta Mater.}, 2019, \textbf{169},
  225--236\relax
\mciteBstWouldAddEndPuncttrue
\mciteSetBstMidEndSepPunct{\mcitedefaultmidpunct}
{\mcitedefaultendpunct}{\mcitedefaultseppunct}\relax
\EndOfBibitem
\bibitem[Rao \emph{et~al.}(2022)Rao, Tung, Xie, Wei, Zhang, Ferrari, Klaver,
  Körmann, Sukumar, da~Silva, Chen, Li, Ponge, Neugebauer, Gutfleisch, Bauer,
  and Raabe]{Ziyuan2022}
Z.~Rao, P.-Y. Tung, R.~Xie, Y.~Wei, H.~Zhang, A.~Ferrari, T.~Klaver,
  F.~Körmann, P.~T. Sukumar, A.~K. da~Silva, Y.~Chen, Z.~Li, D.~Ponge,
  J.~Neugebauer, O.~Gutfleisch, S.~Bauer and D.~Raabe, \emph{Science}, 2022,
  \textbf{378}, 78--85\relax
\mciteBstWouldAddEndPuncttrue
\mciteSetBstMidEndSepPunct{\mcitedefaultmidpunct}
{\mcitedefaultendpunct}{\mcitedefaultseppunct}\relax
\EndOfBibitem
\bibitem[Roberts \emph{et~al.}(2024)Roberts, Rijal, Divilov, Maria,
  Fahrenholtz, Wolfe, Brenner, Curtarolo, and Zurek]{Josiah2024}
J.~Roberts, B.~Rijal, S.~Divilov, J.-P. Maria, W.~G. Fahrenholtz, D.~E. Wolfe,
  D.~W. Brenner, S.~Curtarolo and E.~Zurek, \emph{npj Computational Materials},
  2024, \textbf{10}, 142\relax
\mciteBstWouldAddEndPuncttrue
\mciteSetBstMidEndSepPunct{\mcitedefaultmidpunct}
{\mcitedefaultendpunct}{\mcitedefaultseppunct}\relax
\EndOfBibitem
\bibitem[Alman(2013)]{Alman2013}
D.~Alman, \emph{Entropy}, 2013, \textbf{15}, 4504--4519\relax
\mciteBstWouldAddEndPuncttrue
\mciteSetBstMidEndSepPunct{\mcitedefaultmidpunct}
{\mcitedefaultendpunct}{\mcitedefaultseppunct}\relax
\EndOfBibitem
\bibitem[Curtarolo \emph{et~al.}(2012)Curtarolo, Setyawan, Hart, Jahnatek,
  Chepulskii, Taylor, Wang, Xue, Yang, Levy, Mehl, Stokes, Demchenko, and
  Morgan]{Curtarolo2012}
S.~Curtarolo, W.~Setyawan, G.~L. Hart, M.~Jahnatek, R.~V. Chepulskii, R.~H.
  Taylor, S.~Wang, J.~Xue, K.~Yang, O.~Levy, M.~J. Mehl, H.~T. Stokes, D.~O.
  Demchenko and D.~Morgan, \emph{Computational Materials Science}, 2012,
  \textbf{58}, 218--226\relax
\mciteBstWouldAddEndPuncttrue
\mciteSetBstMidEndSepPunct{\mcitedefaultmidpunct}
{\mcitedefaultendpunct}{\mcitedefaultseppunct}\relax
\EndOfBibitem
\bibitem[Zhang \emph{et~al.}(2014)Zhang, Zhang, Chen, Zhu, Cao, and
  Kattner]{ZHANG20141}
F.~Zhang, C.~Zhang, S.~Chen, J.~Zhu, W.~Cao and U.~Kattner, \emph{Calphad},
  2014, \textbf{45}, 1 -- 10\relax
\mciteBstWouldAddEndPuncttrue
\mciteSetBstMidEndSepPunct{\mcitedefaultmidpunct}
{\mcitedefaultendpunct}{\mcitedefaultseppunct}\relax
\EndOfBibitem
\bibitem[Toher and Curtarolo(2024)]{Toher2024}
C.~Toher and S.~Curtarolo, \emph{Journal of Phase Equilibria and Diffusion},
  2024, \textbf{45}, 219--227\relax
\mciteBstWouldAddEndPuncttrue
\mciteSetBstMidEndSepPunct{\mcitedefaultmidpunct}
{\mcitedefaultendpunct}{\mcitedefaultseppunct}\relax
\EndOfBibitem
\bibitem[Hart \emph{et~al.}(2021)Hart, Mueller, Toher, and Curtarolo]{Gus2021}
G.~L.~W. Hart, T.~Mueller, C.~Toher and S.~Curtarolo, \emph{Nature Reviews
  Materials}, 2021, \textbf{6}, 730--755\relax
\mciteBstWouldAddEndPuncttrue
\mciteSetBstMidEndSepPunct{\mcitedefaultmidpunct}
{\mcitedefaultendpunct}{\mcitedefaultseppunct}\relax
\EndOfBibitem
\bibitem[Nguyen \emph{et~al.}(2023)Nguyen, Kino, Miyake, and Dam]{Nguyen2023}
D.-N. Nguyen, H.~Kino, T.~Miyake and H.-C. Dam, \emph{MRS Bulletin}, 2023,
  \textbf{48}, 31--44\relax
\mciteBstWouldAddEndPuncttrue
\mciteSetBstMidEndSepPunct{\mcitedefaultmidpunct}
{\mcitedefaultendpunct}{\mcitedefaultseppunct}\relax
\EndOfBibitem
\bibitem[Ha \emph{et~al.}(2021)Ha, Nguyen, Nguyen, Nagata, Chikyow, Kino,
  Miyake, Den{\oe}ux, Huynh, and Dam]{Ha2021}
M.-Q. Ha, D.-N. Nguyen, V.-C. Nguyen, T.~Nagata, T.~Chikyow, H.~Kino,
  T.~Miyake, T.~Den{\oe}ux, V.-N. Huynh and H.-C. Dam, \emph{Nature
  Computational Science}, 2021, \textbf{1}, 470--478\relax
\mciteBstWouldAddEndPuncttrue
\mciteSetBstMidEndSepPunct{\mcitedefaultmidpunct}
{\mcitedefaultendpunct}{\mcitedefaultseppunct}\relax
\EndOfBibitem
\bibitem[H{\"u}llermeier and Waegeman(2021)]{Hullermeier2021}
E.~H{\"u}llermeier and W.~Waegeman, \emph{Machine Learning}, 2021,
  \textbf{110}, 457--506\relax
\mciteBstWouldAddEndPuncttrue
\mciteSetBstMidEndSepPunct{\mcitedefaultmidpunct}
{\mcitedefaultendpunct}{\mcitedefaultseppunct}\relax
\EndOfBibitem
\bibitem[Hüllermeier and Brinker(2008)]{Hullermeier2008}
E.~Hüllermeier and K.~Brinker, \emph{Fuzzy Sets and Systems}, 2008,
  \textbf{159}, 2337--2352\relax
\mciteBstWouldAddEndPuncttrue
\mciteSetBstMidEndSepPunct{\mcitedefaultmidpunct}
{\mcitedefaultendpunct}{\mcitedefaultseppunct}\relax
\EndOfBibitem
\bibitem[George \emph{et~al.}(2019)George, Raabe, and Ritchie]{George2019}
E.~P. George, D.~Raabe and R.~O. Ritchie, \emph{Nat. Rev. Mater.}, 2019,
  \textbf{4}, 515--534\relax
\mciteBstWouldAddEndPuncttrue
\mciteSetBstMidEndSepPunct{\mcitedefaultmidpunct}
{\mcitedefaultendpunct}{\mcitedefaultseppunct}\relax
\EndOfBibitem
\bibitem[Konno \emph{et~al.}(2021)Konno, Kurokawa, Nabeshima, Sakishita, Ogawa,
  Hosako, and Maeda]{Konno2021}
T.~Konno, H.~Kurokawa, F.~Nabeshima, Y.~Sakishita, R.~Ogawa, I.~Hosako and
  A.~Maeda, \emph{Phys. Rev. B}, 2021, \textbf{103}, 014509\relax
\mciteBstWouldAddEndPuncttrue
\mciteSetBstMidEndSepPunct{\mcitedefaultmidpunct}
{\mcitedefaultendpunct}{\mcitedefaultseppunct}\relax
\EndOfBibitem
\bibitem[Dempster(1968)]{dempster1968}
A.~P. Dempster, \emph{Journal of the Royal Statistical Society: Series B
  (Methodological)}, 1968, \textbf{30}, 205--232\relax
\mciteBstWouldAddEndPuncttrue
\mciteSetBstMidEndSepPunct{\mcitedefaultmidpunct}
{\mcitedefaultendpunct}{\mcitedefaultseppunct}\relax
\EndOfBibitem
\bibitem[Shafer(1976)]{Shafer1976}
G.~Shafer, \emph{A Mathematical Theory of Evidence}, Princeton University
  Press, 1976\relax
\mciteBstWouldAddEndPuncttrue
\mciteSetBstMidEndSepPunct{\mcitedefaultmidpunct}
{\mcitedefaultendpunct}{\mcitedefaultseppunct}\relax
\EndOfBibitem
\bibitem[Den{\oe}ux \emph{et~al.}(2020)Den{\oe}ux, Dubois, and
  Prade]{denoeux20b}
T.~Den{\oe}ux, D.~Dubois and H.~Prade, \emph{A Guided Tour of Artificial
  Intelligence Research}, Springer Verlag, 2020, vol.~1, ch.~4, pp.
  119--150\relax
\mciteBstWouldAddEndPuncttrue
\mciteSetBstMidEndSepPunct{\mcitedefaultmidpunct}
{\mcitedefaultendpunct}{\mcitedefaultseppunct}\relax
\EndOfBibitem
\bibitem[Ha \emph{et~al.}(2023)Ha, Nguyen, Nguyen, Kino, Ando, Miyake,
  Den{\oe}ux, Huynh, and Dam]{Ha2023}
M.-Q. Ha, D.-N. Nguyen, V.-C. Nguyen, H.~Kino, Y.~Ando, T.~Miyake,
  T.~Den{\oe}ux, V.-N. Huynh and H.-C. Dam, \emph{Journal of Applied Physics},
  2023, \textbf{133}, 053904\relax
\mciteBstWouldAddEndPuncttrue
\mciteSetBstMidEndSepPunct{\mcitedefaultmidpunct}
{\mcitedefaultendpunct}{\mcitedefaultseppunct}\relax
\EndOfBibitem
\bibitem[Nu~Thanh~Ton \emph{et~al.}(2020)Nu~Thanh~Ton, Ha, Ikenaga, Thakur,
  Dam, and Taniike]{Ton2021}
N.~Nu~Thanh~Ton, M.-Q. Ha, T.~Ikenaga, A.~Thakur, H.-C. Dam and T.~Taniike,
  \emph{2D Materials}, 2020, \textbf{8}, 015019\relax
\mciteBstWouldAddEndPuncttrue
\mciteSetBstMidEndSepPunct{\mcitedefaultmidpunct}
{\mcitedefaultendpunct}{\mcitedefaultseppunct}\relax
\EndOfBibitem
\bibitem[Pyzer-Knapp \emph{et~al.}(2022)Pyzer-Knapp, Pitera, Staar, Takeda,
  Laino, Sanders, Sexton, Smith, and Curioni]{Knapp2022}
E.~O. Pyzer-Knapp, J.~W. Pitera, P.~W.~J. Staar, S.~Takeda, T.~Laino, D.~P.
  Sanders, J.~Sexton, J.~R. Smith and A.~Curioni, \emph{npj Computational
  Materials}, 2022, \textbf{8}, 84\relax
\mciteBstWouldAddEndPuncttrue
\mciteSetBstMidEndSepPunct{\mcitedefaultmidpunct}
{\mcitedefaultendpunct}{\mcitedefaultseppunct}\relax
\EndOfBibitem
\bibitem[Cook \emph{et~al.}(2024)Cook, Kumar, Payne, Belcher, Borges, Wang,
  Walsh, Li, Devaraj, Zhang, Asta, Minor, Lavernia, Apelian, and
  Ritchie]{David2024}
D.~H. Cook, P.~Kumar, M.~I. Payne, C.~H. Belcher, P.~Borges, W.~Wang, F.~Walsh,
  Z.~Li, A.~Devaraj, M.~Zhang, M.~Asta, A.~M. Minor, E.~J. Lavernia, D.~Apelian
  and R.~O. Ritchie, \emph{Science}, 2024, \textbf{384}, 178--184\relax
\mciteBstWouldAddEndPuncttrue
\mciteSetBstMidEndSepPunct{\mcitedefaultmidpunct}
{\mcitedefaultendpunct}{\mcitedefaultseppunct}\relax
\EndOfBibitem
\bibitem[Liu \emph{et~al.}(2024)Liu, Wen, Pattamatta, and Srolovitz]{Liu2024}
S.~Liu, T.~Wen, A.~S. Pattamatta and D.~J. Srolovitz, \emph{Materials Today},
  2024, \textbf{80}, 240--249\relax
\mciteBstWouldAddEndPuncttrue
\mciteSetBstMidEndSepPunct{\mcitedefaultmidpunct}
{\mcitedefaultendpunct}{\mcitedefaultseppunct}\relax
\EndOfBibitem
\bibitem[Tversky(1977)]{Tversky1978}
A.~Tversky, \emph{Psychological Review}, 1977, \textbf{84}, 327--352\relax
\mciteBstWouldAddEndPuncttrue
\mciteSetBstMidEndSepPunct{\mcitedefaultmidpunct}
{\mcitedefaultendpunct}{\mcitedefaultseppunct}\relax
\EndOfBibitem
\bibitem[Smets(1993)]{SMETS1993}
P.~Smets, \emph{International Journal of Approximate Reasoning}, 1993,
  \textbf{9}, 1--35\relax
\mciteBstWouldAddEndPuncttrue
\mciteSetBstMidEndSepPunct{\mcitedefaultmidpunct}
{\mcitedefaultendpunct}{\mcitedefaultseppunct}\relax
\EndOfBibitem
\bibitem[Chen \emph{et~al.}(2023)Chen, Hilhorst, Bokas, Gorsse, Jacques, and
  Hautier]{chen2023}
W.~Chen, A.~Hilhorst, G.~Bokas, S.~Gorsse, P.~J. Jacques and G.~Hautier,
  \emph{Nature Communications}, 2023, \textbf{14}, 2856\relax
\mciteBstWouldAddEndPuncttrue
\mciteSetBstMidEndSepPunct{\mcitedefaultmidpunct}
{\mcitedefaultendpunct}{\mcitedefaultseppunct}\relax
\EndOfBibitem
\bibitem[Takeuchi and Inoue(2005)]{Takeuchi2005}
A.~Takeuchi and A.~Inoue, \emph{MATERIALS TRANSACTIONS}, 2005, \textbf{46},
  2817--2829\relax
\mciteBstWouldAddEndPuncttrue
\mciteSetBstMidEndSepPunct{\mcitedefaultmidpunct}
{\mcitedefaultendpunct}{\mcitedefaultseppunct}\relax
\EndOfBibitem
\bibitem[Takeuchi and Inoue(2010)]{Takeuchi2010}
A.~Takeuchi and A.~Inoue, \emph{Intermetallics}, 2010, \textbf{18},
  1779--1789\relax
\mciteBstWouldAddEndPuncttrue
\mciteSetBstMidEndSepPunct{\mcitedefaultmidpunct}
{\mcitedefaultendpunct}{\mcitedefaultseppunct}\relax
\EndOfBibitem
\bibitem[Fukushima \emph{et~al.}(2022)Fukushima, Akai, Chikyow, and
  Kino]{Fukushima2022}
T.~Fukushima, H.~Akai, T.~Chikyow and H.~Kino, \emph{Phys. Rev. Materials},
  2022, \textbf{6}, 023802\relax
\mciteBstWouldAddEndPuncttrue
\mciteSetBstMidEndSepPunct{\mcitedefaultmidpunct}
{\mcitedefaultendpunct}{\mcitedefaultseppunct}\relax
\EndOfBibitem
\bibitem[LaValley(2008)]{Michael2008}
M.~P. LaValley, \emph{Circulation}, 2008, \textbf{117}, 2395--2399\relax
\mciteBstWouldAddEndPuncttrue
\mciteSetBstMidEndSepPunct{\mcitedefaultmidpunct}
{\mcitedefaultendpunct}{\mcitedefaultseppunct}\relax
\EndOfBibitem
\end{mcitethebibliography}


\providecommand*{\mcitethebibliography}{\thebibliography}
\csname @ifundefined\endcsname{endmcitethebibliography}
{\let\endmcitethebibliography\endthebibliography}{}
\begin{mcitethebibliography}{3}
\providecommand*{\natexlab}[1]{#1}
\providecommand*{\mciteSetBstSublistMode}[1]{}
\providecommand*{\mciteSetBstMaxWidthForm}[2]{}
\providecommand*{\mciteBstWouldAddEndPuncttrue}
  {\def\EndOfBibitem{\unskip.}}
\providecommand*{\mciteBstWouldAddEndPunctfalse}
  {\let\EndOfBibitem\relax}
\providecommand*{\mciteSetBstMidEndSepPunct}[3]{}
\providecommand*{\mciteSetBstSublistLabelBeginEnd}[3]{}
\providecommand*{\EndOfBibitem}{}
\mciteSetBstSublistMode{f}
\mciteSetBstMaxWidthForm{subitem}
{(\emph{\alph{mcitesubitemcount}})}
\mciteSetBstSublistLabelBeginEnd{\mcitemaxwidthsubitemform\space}
{\relax}{\relax}

\bibitem[Dempster(1968)]{dempster1968}
A.~P. Dempster, \emph{Journal of the Royal Statistical Society: Series B
  (Methodological)}, 1968, \textbf{30}, 205--232\relax
\mciteBstWouldAddEndPuncttrue
\mciteSetBstMidEndSepPunct{\mcitedefaultmidpunct}
{\mcitedefaultendpunct}{\mcitedefaultseppunct}\relax
\EndOfBibitem
\bibitem[Seko \emph{et~al.}(2018)Seko, Togo, and Tanaka]{Seko2018}
A.~Seko, A.~Togo and I.~Tanaka, in \emph{Descriptors for Machine Learning of
  Materials Data}, ed. I.~Tanaka, Springer Singapore, Singapore, 2018, pp.
  3--23\relax
\mciteBstWouldAddEndPuncttrue
\mciteSetBstMidEndSepPunct{\mcitedefaultmidpunct}
{\mcitedefaultendpunct}{\mcitedefaultseppunct}\relax
\EndOfBibitem
\bibitem[Seko \emph{et~al.}(2017)Seko, Hayashi, Nakayama, Takahashi, and
  Tanaka]{Seko2017}
A.~Seko, H.~Hayashi, K.~Nakayama, A.~Takahashi and I.~Tanaka, \emph{Phys. Rev.
  B}, 2017, \textbf{95}, 144110\relax
\mciteBstWouldAddEndPuncttrue
\mciteSetBstMidEndSepPunct{\mcitedefaultmidpunct}
{\mcitedefaultendpunct}{\mcitedefaultseppunct}\relax
\EndOfBibitem
\end{mcitethebibliography}
\bibliographystyle{rsc} 
\end{document}


\preprint{AIP/123-QED}

\title[Supplementary Information]{Supplementary Information}

\author{Minh-Quyet Ha}
\affiliation{Japan Advanced Institute of Science and Technology, 1-1 Asahidai, Nomi, Ishikawa 923-1292, Japan}
 
 \author{Dinh-Khiet Le}
\affiliation{Japan Advanced Institute of Science and Technology, 1-1 Asahidai, Nomi, Ishikawa 923-1292, Japan}

 \author{Duc-Anh Dao}
\affiliation{Japan Advanced Institute of Science and Technology, 1-1 Asahidai, Nomi, Ishikawa 923-1292, Japan}

 \author{Tien-Sinh Vu}
\affiliation{Japan Advanced Institute of Science and Technology, 1-1 Asahidai, Nomi, Ishikawa 923-1292, Japan}
 
\author{Duong-Nguyen Nguyen}%
\affiliation{Japan Advanced Institute of Science and Technology, 1-1 Asahidai, Nomi, Ishikawa 923-1292, Japan}

\author{Viet-Cuong Nguyen}
\affiliation{%
HPC SYSTEMS Inc., Minato, Tokyo 108-0022, Japan
}

%
 
\author{Hiori Kino}
 \affiliation{MaDIS, National Institute for Materials Science, 1-2-1 Sengen, Tsukuba, Ibaraki 305-0044, Japan}


\author{Thierry Denœux}
 \affiliation{Université de technologie de Compiègne, CNRS, UMR 7253 Heudiasyc, Compiègne, France}

\author{Van-Nam Huynh}
 \affiliation{Japan Advanced Institute of Science and Technology, 1-1 Asahidai, Nomi, Ishikawa 923-1292, Japan}

\author{Hieu-Chi Dam}
\email{dam@jaist.ac.jp}
\affiliation{Japan Advanced Institute of Science and Technology, 1-1 Asahidai, Nomi, Ishikawa 923-1292, Japan}

\date{\today}

\maketitle

\section{\label{sec:example}{Illustrative examples}}

The following examples provide explanations of how the evidence theory work to learn the similarity and infer the HEA formation for new element combinations, identifying equiatomic alloys.

\textbf{Example 1:} Suppose we have collected four pairs of alloys from experiments. Three of those pairs are alloys that both form HEA phase: $pair_1=(\{A^1,B^1,C^1,D\}, \{A^1,B^1,C^1,E\})$; $pair_2=(\{A^2,B^2,C^2,D\},\{A^2,B^2,C^2,E\})$; and $pair_3=(\{A^3,B^3,C^3,D\}, \{A^3,B^3,C^3,E\})$.  The fourth pair $pair_4=(\{A^4,B^4,C^4,D\},\{A^4,B^4,C^4,E\})$ is different from the other three, in which $\{A^4,B^4,C^4,D\}$ forms HEA phase while $\{A^4,B^4,C^4,E\}$ does not form HEA phase. We consider each pair as a source of evidence support that $\{D\}$ is similar to $\{E\}$ in term of substitutability to form the HEA phase. Each evidence is modeled using mass function as follows:

\begin{equation*}
\begin{aligned}
&m^{\{C\},\{D\}}_{pair_1}(\{similar\})=0.1,\\
&m^{\{C\},\{D\}}_{pair_1}(\{dissimilar\})=0,\\
&m^{\{C\},\{D\}}_{pair_1}(\{similar, dissimilar\})=0.9
\end{aligned}
\end{equation*}

\begin{equation*}
\begin{aligned}
&m^{\{C\},\{D\}}_{pair_2}(\{similar\})=0.1,\\
&m^{\{C\},\{D\}}_{pair_2}(\{dissimilar\})=0,\\
&m^{\{C\},\{D\}}_{pair_2}(\{similar, dissimilar\})=0.9
\end{aligned}
\end{equation*}

\begin{equation*}
\begin{aligned}
&m^{\{C\},\{D\}}_{pair_3}(\{similar\})=0.1,\\
&m^{\{C\},\{D\}}_{pair_3}(\{dissimilar\})=0,\\
&m^{\{C\},\{D\}}_{pair_3}(\{similar, dissimilar\})=0.9
\end{aligned}
\end{equation*}

\begin{equation*}
\begin{aligned}
&m^{\{C\},\{D\}}_{pair_4}(\{similar\})=0,\\
&m^{\{C\},\{D\}}_{pair_4}(\{dissimilar\})=0.1,\\
&m^{\{C\},\{D\}}_{pair_4}(\{similar, dissimilar\})=0.9
\end{aligned}
\end{equation*}

The three pieces of evidence are combined using the Dempster' rule of combination to accumulate the believe that $\{D\}$ is similar to $\{E\}$:

\begin{equation*}
\begin{aligned}
&m^{\{C\},\{D\}}(\{similar\})=0.25,\\
&m^{\{C\},\{D\}}(\{dissimilar\})=0.075,\\
&m^{\{C\},\{D\}}(\{similar, dissimilar\})=0.675
\end{aligned}
\end{equation*}

Next, if we observed (included in the data) that the HEA phase exists for alloy $\{G,H,I,D\}$, the ERS (which focuses on finding some chance for discovering new combination of elements that the HEA phase exist and ignores the belief regarding $\neg HEA$) will consider that there is some believe that the HEA phase also exists for $\{G,H,I,E\}$ (by substituting $\{D\}$ with $\{E\}$). The evidence is modeled using mass function as follows:

\begin{equation*}
\begin{aligned}
&m^{\{G,H,I,E\}}_{\{G,H,I,D\}, \{D\}\leftarrow \{E\}}(\{\neg HEA\})=0,\\
&m^{\{G,H,I,E\}}_{\{G,H,I,D\}, \{D\}\leftarrow \{E\}}(\{HEA\})=m^{C,D}(\{similar\})=0.25,\\
&m^{\{G,H,I,E\}}_{\{G,H,I,D\}, \{D\}\leftarrow \{E\}}(\{HEA, \neg HEA\})=1-m^{C,D}(\{similar\})=0.75
\end{aligned}
\end{equation*}

\textbf{Example 2:} In a same manner but for an extrapolative recommendation: if the HEA phases exist for all the alloys in the three following pairs: $pair_1=(\{A^1,B^1,C\}, \{A^1,B^1,D,E\})$, $pair_2=(\{A^2,B^2,C\}, \{A^2,B^2,D,E\})$, $pair_3=(\{A^3,B^3,C\}, \{A^3,B^3,D,E\})$.  In the fourth pair $pair_4=(\{A^4,B^4,C\},\{A^4,B^4,D,E\})$,  $\{A^4,B^4,C\}$ forms HEA phase while $\{A^4,B^4,D,E\}$ does not form HEA phase. The algorithm will accumulate the believe that $\{C\}$ is similar to $\{D,E\}$ as follows: 

\begin{equation*}
\begin{aligned}
&m^{\{C\},\{D,E\}}(\{similar\})=0.25,\\
&m^{\{C\},\{D,E\}}(\{dissimilar\})=0.075,\\
&m^{\{C\},\{D,E\}}(\{similar, dissimilar\})=0.675
\end{aligned}
\end{equation*}

Consequently, if we observed (included in the data) that the HEA phase exists for $\{G,H,I,C\}$, the algorithm (which focuses on finding some chance for discovering new combination of elements that the HEA phase exist and ignores the belief regarding $\neg HEA$) will consider that there is some believe that the HEA phase also exists for $\{G,H,I,D,E\}$ (by substituting $\{C\}$ with $\{D,E\}$).

\begin{equation*}
\begin{aligned}
&m^{\{G,H,I,D,E\}}_{\{G,H,I,C\}, \{C\}\leftarrow \{D,E\}}(\{\neg HEA\})=0,\\
&m^{\{G,H,I,D,E\}}_{\{G,H,I,C\}, \{C\}\leftarrow \{D,E\}}(\{HEA\})=m^{C,D}(\{similar\})=0.25,\\
&m^{\{G,H,I,D,E\}}_{\{G,H,I,C\}, \{C\}\leftarrow \{D,E\}}(\{HEA, \neg HEA\})=1-m^{C,D}(\{similar\})=0.75
\end{aligned}
\end{equation*}

 \section{\label{subsec:dempster_rule}{Combining pieces of evidence using Dempster's rule of combination}}
 
We assume that we can collect $q$ pieces of evidence from $\mathcal{D}$ to compare a specific pair of element combinations, $C_t$ and $C_v$. If no evidence is found, the mass function $m^{C_t,C_v}_{none}$ is initialized, which assigns a probability mass of 1 to subset $\{similar, dissimilar\}$. $m^{C_t,C_v}_{none}$ models the condition under which no information about the similarity (or dissimilarity) between $C_t$ and $C_v$ is available. Any two pieces of evidence $a$ and $b$ modeled by the corresponding mass functions $m_a^{C_t,C_v}$ and $m_b^{C_t,C_v}$ can be combined using the Dempster rule \cite{dempster1968} to assign the joint mass $m_{a,b}^{C_t,C_v}$ to each subset $\omega$ of $\Omega_{sim}$ (i.e. $\{similar\}$, $\{dissimilar\}$, or $\{similar, dissimilar\}$) as follows:
\begin{equation}
\begin{aligned}
m_{a,b}^{C_t,C_v}(\omega)={}& \left(m_a^{C_t,C_v} \oplus m_b^{C_t,C_v} \right)(\omega) \\
={}& { \frac{\sum\limits_{\forall\omega_k \cap \omega_h = \omega} {m_a^{C_t,C_v}(\omega_k) \times m_b^{C_t,C_v}(\omega_h) }}{1 - \sum\limits_{\forall\omega_k \cap \omega_h = \emptyset} {m_a^{C_t,C_v}(\omega_k) \times m_b^{C_t,C_v}(\omega_h)}}},
\end{aligned}
\end{equation}
where $\omega$, $\omega_k$ and $\omega_h$ are subsets of $\Omega_{sim}$. Note that the Dempster rule is commutative and yields the same result by changing the order of $m_a^{C_t,C_v}$ and $m_b^{C_t,C_v}$. All the $q$ obtained mass functions corresponding to the $q$ collected pieces of evidence from available sources are then combined using the Dempster rule to assign the final mass $m^{C_t,C_v}$ as follows: 
\begin{equation}
m^{C_t,C_v}_{\mathcal{D}}(\omega) = \left(m_1^{C_t,C_v} \oplus m_2^{C_t,C_v} \oplus \dots \oplus m_q^{C_t,C_v} \right)(\omega).
\end{equation} 

 
 \section{\label{subsec:descriptors}{Materials descriptors}}

Descriptors, which are the representation of alloys, play a crucial role in building a recommender system to explore potential new HEAs. In this research, the raw data of alloys is represented in the form of elements combination. Several descriptors have been studied in materials informatics to represent the compounds \cite{Seko2018}. To employ the data-driven approaches for this work, we applied compositional descriptor\cite{Seko2017}. 

Compositional descriptor represents an alloy by a set of 135 features composed of means, standard deviations, and covariance of established atomic representations that form the alloy. The descriptor can be applied not only to crystalline systems but also to molecular system. We adopted 15 atomic representations: (1) atomic number, (2) atomic mass,(3) period and (4) group in the periodic table, (5) first ionization energy, (6) second ionization energy, (7) Pauling electronegativity, (8) Allen electronegativity, (9) van der Waals radius, (10) covalentradius, (11) atomic radius, (12) melting point, (13) boiling point, (14) density, and (15) specific heat. However, the compositional descriptor hardly distinguishes compounds which have different numbers of the atom because it is to regard the atomic representations of a compound as distributions of data. Therefore, the compositional descriptor cannot be applied in the case of having extrapolation in the number of components.

\section{\label{subsec:alpha_estimation}Tuning hyper-parameter of the predictive models}

Because the datasets used in this study are derived from calculation-based prediction methods, we introduce a degree of uncertainty $\alpha$ into the mass function that models substitutability evidence from material datasets. For each dataset, a grid search is performed to determine the optimal $\alpha$, which best reproduces the alloy labels by achieving the highest cross-validation score. The cross-validation scheme employs a 10-fold cross-validation repeated three times. The search space for $\alpha$ ranges from 0.01 to 0.5, with increments of 0.01.

In contrast, the degree of uncertainty $\beta$ in the mass function representing substitutability evidence from domain knowledge is assigned a fixed value of $1/N_{domains}$, where $N_{domains}$ denotes the number of knowledge domains applied in the study. This ensures balanced weighting of evidence across multiple sources.

Additionally, the grid search parameter tuning is also applied to other traditional models. The parameter grid for logistic regression is as follows:
\begin{itemize}
\item \textbf{Penalty}: {\texttt{l1}, \texttt{l2}}
\item \textbf{Solver}: {
\begin{itemize}
\item \texttt{l1} penalty: [\texttt{liblinear}, \texttt{saga}]
\item \texttt{l2} penalty:
\begin{itemize}
\item \texttt{dual=True}: [\texttt{liblinear}]
\item \texttt{dual=False}: [\texttt{newton-cg}, \texttt{lbfgs}, \texttt{sag}, \texttt{saga}, \texttt{liblinear}]
\end{itemize}
\end{itemize}
}
\item \textbf{Regularization Parameter} $C$: [0.01, 0.1, 1, 10, 100]
\end{itemize}

\section{Hybrid distance measurement for alloys}\label{sec.ers_jaccard_distance_matrix}

To identify the subgroup to which unobserved alloys belong, we propose an approach that constructs an embedding space for both observed and unobserved alloys based on a combined distance matrix $\overline{M}$. This matrix is computed as the element-wise product of the distance matrix obtained from our proposed substitutability-based similarity measurement and the distance matrix derived using the Jaccard index for the binary descriptors of the alloys.

For any pair of alloys $A_i$ and $A_j$, the combined distance is defined as follows:
\begin{equation}
\overline{M}[i,j] = \left(m^{C_t,C_v}(\{\mathrm{dissimilar}\}) + \frac{m^{C_t,C_v}(\{\mathrm{similar}, \mathrm{dissimilar}\})}{2}\right) \times (1 - J(A_i, A_j)).
\end{equation}

Here, $m^{C_t,C_v}$ represents the substitutability between the two element combinations, $C_t = A_i \setminus (A_i \cap A_j) \quad \text{and} \quad C_v = A_j \setminus (A_i \cap A_j)$, obtained through our method, while $J(A_i, A_j)$ denotes the Jaccard similarity coefficient between the element sets of the two alloys.

If one of the alloys in the pair is unobserved in the dataset, our method assigns no similarity information between the alloys, resulting in $m^{C_t,C_v}(\{\mathrm{similar}, \mathrm{dissimilar}\})= 1$ and $m^{C_t,C_v}(\{\mathrm{dissimilar}\}) = 0$. In this case, the combined distance simplifies to:
\begin{equation}
\overline{M}[i,j] = \frac{1 - J(A_i, A_j)}{2}
\end{equation}

This approach allows for the effective grouping of unobserved alloys by leveraging both compositional similarity and substitutability information.

\clearpage

\begin{table*}[t]
\centering
\caption{\label{tab:classification-results-Os}Classification report of our proposed framework for extrapolation experiment of Os-based alloys on the dataset $\mathcal{D}_{0.9T_m}$.}
\begin{tabular*}{\textwidth}{@{\extracolsep{\fill}}lcccc}
 &   \textbf{precision}  & \textbf{recall} & \textbf{f1-score} & \textbf{support} \\
\hline
Single-phase & $0.90$ & $0.78$ & $0.84$ & $882$ alloys \\
Multi-phase & $0.88$ & $0.95$ & $0.91$ & $1,418$ alloys \\
 &  &  & &  \\
Accuracy &  &  & $0.88$ & $2,300$ alloys \\
Macro average& $0.89$ & $0.87$ & $0.87$ & $2,300$ alloys \\
Micro average & $0.89$ & $0.88$ & $0.88$ & $2,300$ alloys \\
\end{tabular*}
\end{table*}

\clearpage
\begin{figure}[t]
\centering
  \includegraphics[width=\linewidth]{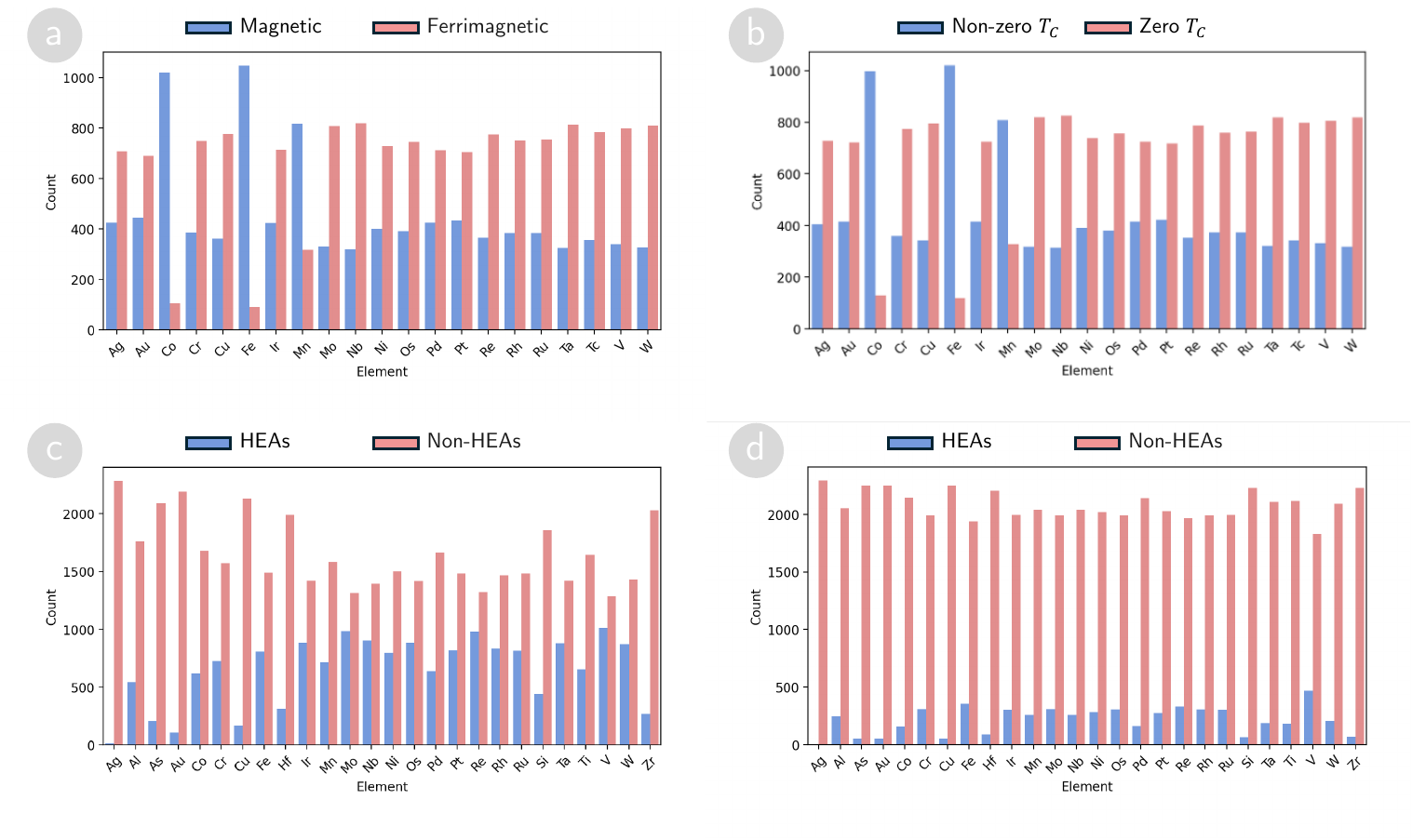}
\caption{\label{fig.s1}Proportions of constituent elements in the four alloys datasets $\mathcal{D}_{\text{Mag}}$ (a), $\mathcal{D}_{T_C}$(b), $\mathcal{D}_{\text{0.9}T_{m}}$ (c),  and $\mathcal{D}_{\text{1350K}}$ (d).}
\end{figure}

\clearpage
\begin{figure}[t]
\centering
  \includegraphics[width=\linewidth]{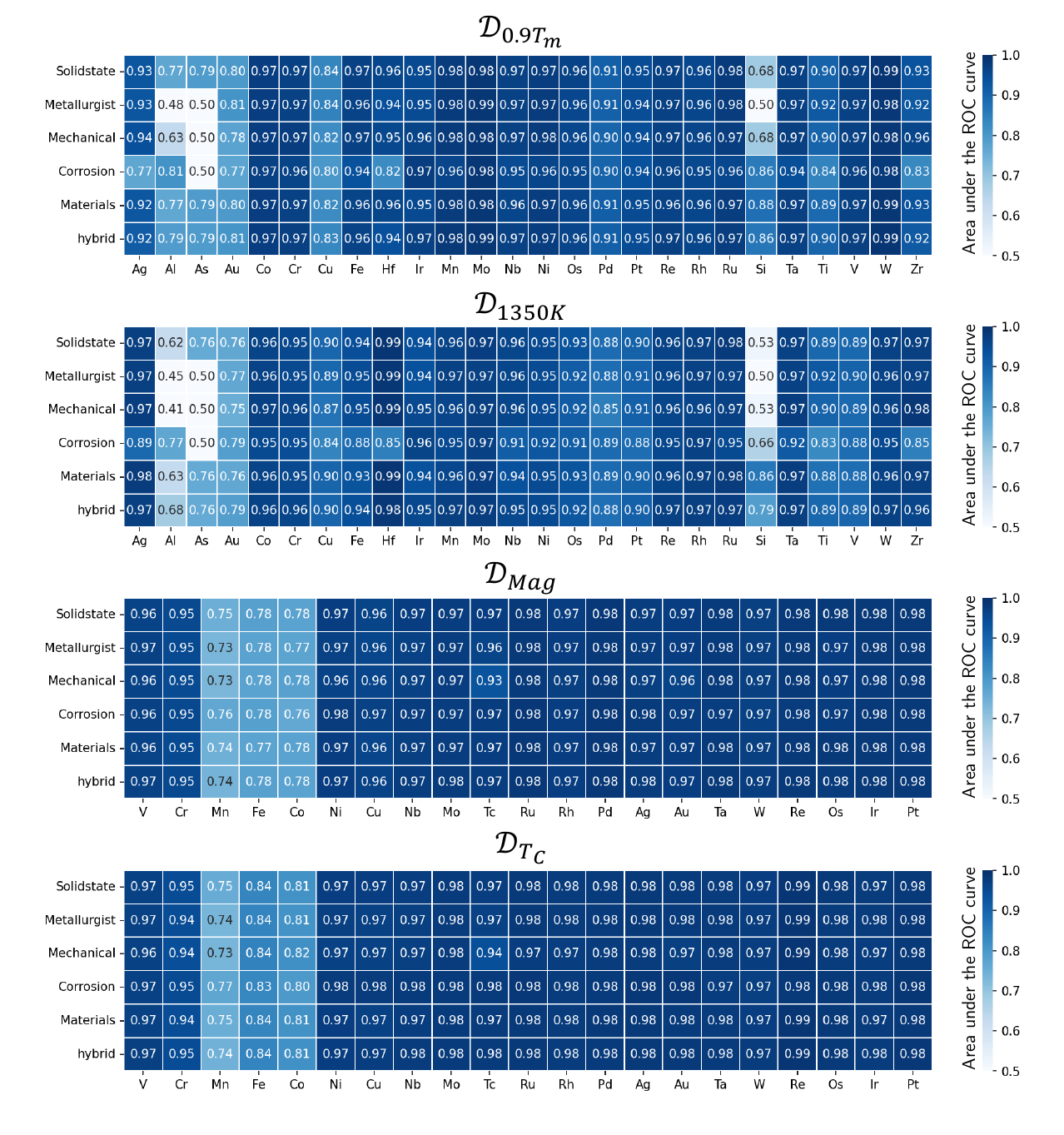}
\caption{\label{fig.s2}The area under the ROC curve (AUC) for extrapolation experiments on constituent elements across the four alloy datasets:  $\mathcal{D}_{\text{Mag}}$, $\mathcal{D}_{T_C}$, $\mathcal{D}_{\text{0.9}T_{m}}$,  and $\mathcal{D}_{\text{1350K}}$.}
\end{figure}

\clearpage
\begin{figure}[t]
\centering
  \includegraphics[width=\linewidth]{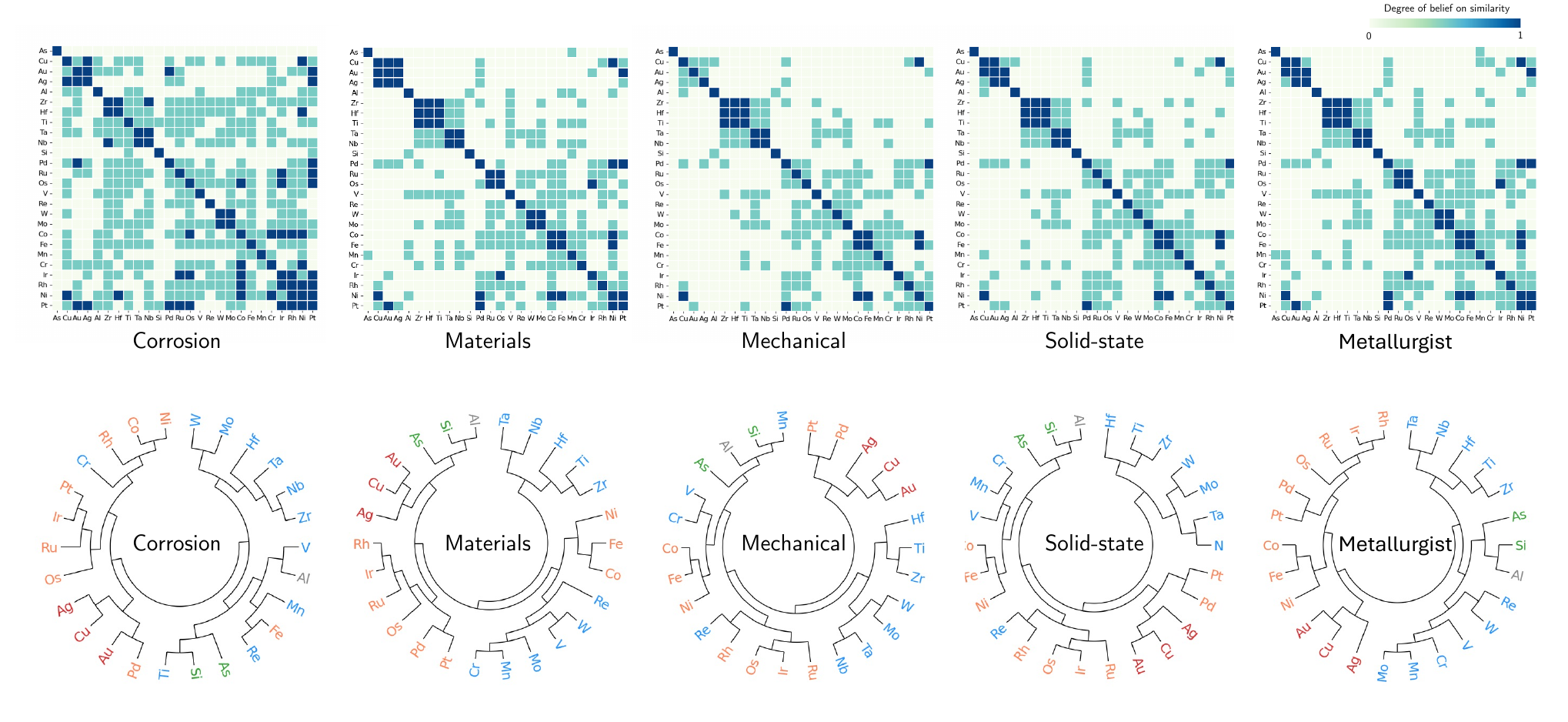}
\caption{\label{fig.s3}(Top) Heatmap illustrating similarity matrices derived from domain knowledge distilled by GPT-4o. The heatmap visualizes the pairwise similarity between elements, based on substitutability evidence extracted from domain-specific knowledge using GPT-4o. The color intensity indicates the degree of similarity, with higher values representing stronger substitutability relationships. (Bottom) Circular hierarchical clustering (HAC) of elements based on substitutability between elements.The circular dendrogram displays the hierarchical clustering of all constituent elements, constructed using hierarchical agglomerative clustering (HAC) with the "complete" linkage criterion. The substitutability information is derived from LLM-based knowledge. Blue labels represent early transition metals, orange labels indicate late transition metals, and red labels denote coinage metals, including copper (Cu), silver (Ag), and gold (Au).}
\end{figure}

\clearpage
\begin{figure}[t]
\centering
  \includegraphics[width=\linewidth]{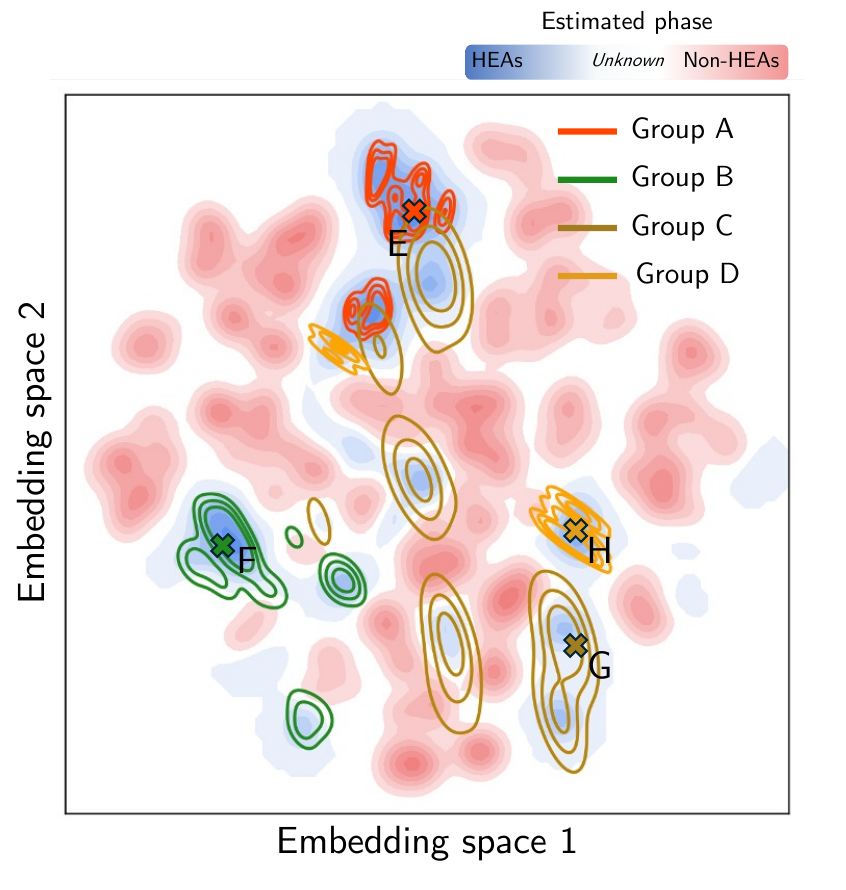}
\caption{\label{fig.s4}Updated map after integrating Os-based alloys into the dataset. The visualization highlights structural changes and reorganization in alloy clusters following the inclusion of Os-based alloys.}
\end{figure}

\clearpage
\begin{figure}[t]
\centering
  \includegraphics[width=\linewidth]{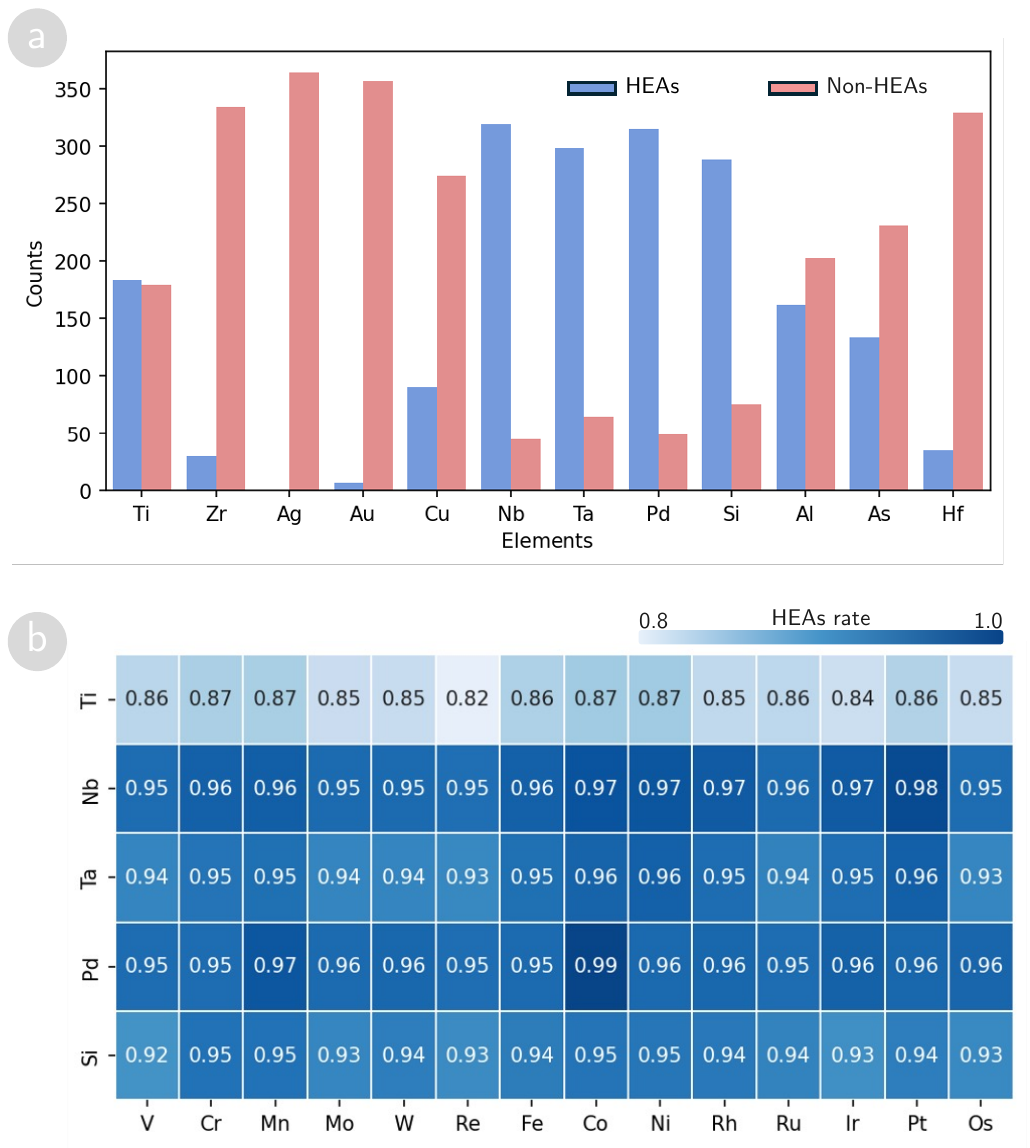}
\caption{\label{fig.s5}Analysis of synthesized effects between transition metals in $\mathcal{E}$ and other elements. (a) Summary of the number of single-phase and multi-phase alloys formed by combining transition metals in $\mathcal{E}$ with other elements. (b) Heatmap showing the stability rate (proportion of single-phase alloys) for alloys formed exclusively from modified sets of $\mathcal{E}$, where one element is replaced by silicon (Si), palladium (Pd), tantalum (Ta), niobium (Nb), or titanium (Ti).}
\end{figure}

\clearpage
\bibliography{supplementary_information.bib} 
\bibliographystyle{rsc} 